%% file: sample-acmtog-SIGGRAPH-submission.tex
\documentclass[acmtog,nonacm]{acmart}
\usepackage[utf8]{inputenc}
\usepackage{newunicodechar}
\newunicodechar{，}{,}
\acmSubmissionID{1214}

\usepackage{booktabs} 

\usepackage[ruled]{algorithm2e} 

\SetAlFnt{\small}
\SetAlCapFnt{\small}
\SetAlCapNameFnt{\small}
\SetAlCapHSkip{0pt}

\acmJournal{TOG}

\newcommand{\eg}[1]{\textit{e.g.}}
\newcommand{\ie}[1]{\textit{i.e.}}

\usepackage[table]{xcolor}
\usepackage{enumitem}
\usepackage{makecell}

\begin{document}
\title{CosmoH2G: A Hand-to-Gripper Transfer Dataset and Baseline Method for Object Manipulation with Complex Spatial Movements}

\author{Hongxiang Zhao}
\orcid{0009-0009-1724-4283}
\email{hongxiangzhao98@gmail.com}
\affiliation{
 \institution{SSE, CUHKSZ}
 \country{China}
}

\author{Mutian Xu$^{\dagger}$}
\orcid{0000-0001-8123-6493}
\email{mutianxu@link.cuhk.edu.cn}
\affiliation{
 \institution{GenuX}
 \country{China}
}

\author{Zeyu Jin}
\orcid{0009-0001-2245-3380}
\email{zeyujin2@link.cuhk.edu.cn}
\affiliation{
 \institution{SSE, CUHKSZ}
 \country{China}
}

\author{Yiming Hao}
\orcid{0009-0000-8185-4400}
\email{haoym1016@gmail.com}
\affiliation{
 \institution{SSE, CUHKSZ}
 \country{China}
}

\author{Shuguang Cui}
\orcid{0000-0003-2608-775X}
\email{shuguangcui@cuhk.edu.cn}
\affiliation{
 \institution{SSE, CUHKSZ;}
 \institution{FNii-Shenzhen}
 \country{China}
}

\author{Xiaoguang Han$^{\dagger}$}
\orcid{0000-0003-0162-3296}
\email{hanxiaoguang@cuhk.edu.cn}
\affiliation{
 \institution{SSE, CUHKSZ;}
 \institution{FNii-Shenzhen;}
 \institution{GenuX}
 \country{China}
}


\begin{abstract}
Transferring human hand demonstrations to robotic grippers has recently emerged as a cost-effective solution for robot learning. However, existing methods are largely confined to simple, planar tasks and fail to handle complex spatial movements (\eg, intricate trajectories involving rotations or flips) that are essential for robot manipulation. 
Motivated by this gap, we adopt an \textit{implicit, data-driven} approach guided by fine-grained \textit{hand-pose} motions.
To this end, we introduce a scalable acquisition pipeline to collect hand-gripper \textit{paired} demonstrations, governed by a rigorous protocol that prioritizes motion complexity and leverages a handheld gripper for seamless action mimicry. This yields a large-scale paired dataset comprising 6,189 episodes across 1,254 unique objects, exhibiting significantly higher spatial complexity than existing benchmarks. However, learning such complex mappings remains challenging. We observe that naive end-to-end generation of full gripper pose sequences is insufficient, as minor trajectory deviations compound rapidly under intricate dynamics. To address this, we propose a two-stage framework: Stage I predicts sparse gripper keyframes (initial and terminal) to simplify the mapping objective, while Stage II generates the full continuous action sequence conditioned on these keyframes. Furthermore, to mitigate cumulative drift, we keep the gripper's orientation being learned while post-optimizing its translation based on the grasping heuristic and kinematic consistency. In both \textit{simulation} and \textit{real-robot} experiments, our framework enables stable and precise hand-to-gripper transfer of complex spatial manipulations, significantly outperforming traditional baselines. Project page: \url{https://cosmoh2g.github.io}.
\end{abstract}



%
%


\begin{teaserfigure}
  \includegraphics[width=0.88\textwidth]{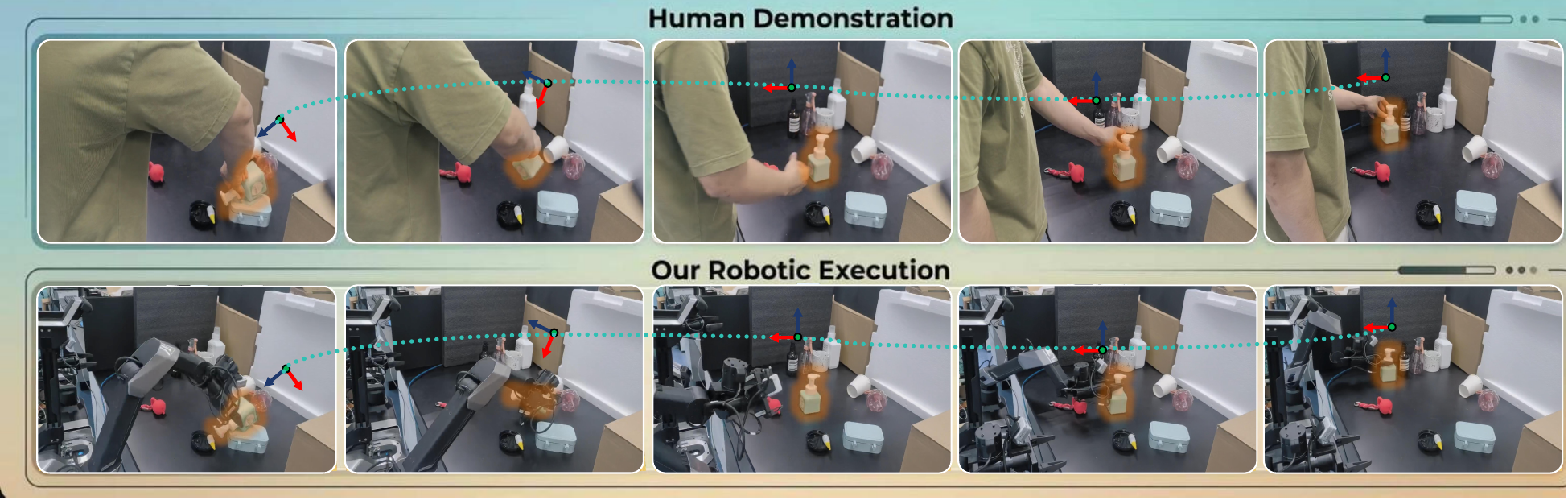}
  \caption{Our CosmoH2G is a data-driven framework that transfers human hand motions to robot gripper actions. Given monocular RGB videos of human demonstrations — including hand rotations and object flips in cluttered scenes — it converts observed hand trajectories into executable gripper motions, supporting stable execution even for intricate spatial movements.}
  \label{fig:teaser}
\end{teaserfigure}

\maketitle

\renewcommand{\thefootnote}{\fnsymbol{footnote}}

\begingroup
\footnotetext[2]{Corresponding Author.}
\endgroup

\input{samplebody-journals}

\end{document}

%% file: samplebody-journals.tex
\section{Introduction}

\begin{figure}[htbp]
  \centering             
  \includegraphics[width=\linewidth, keepaspectratio]{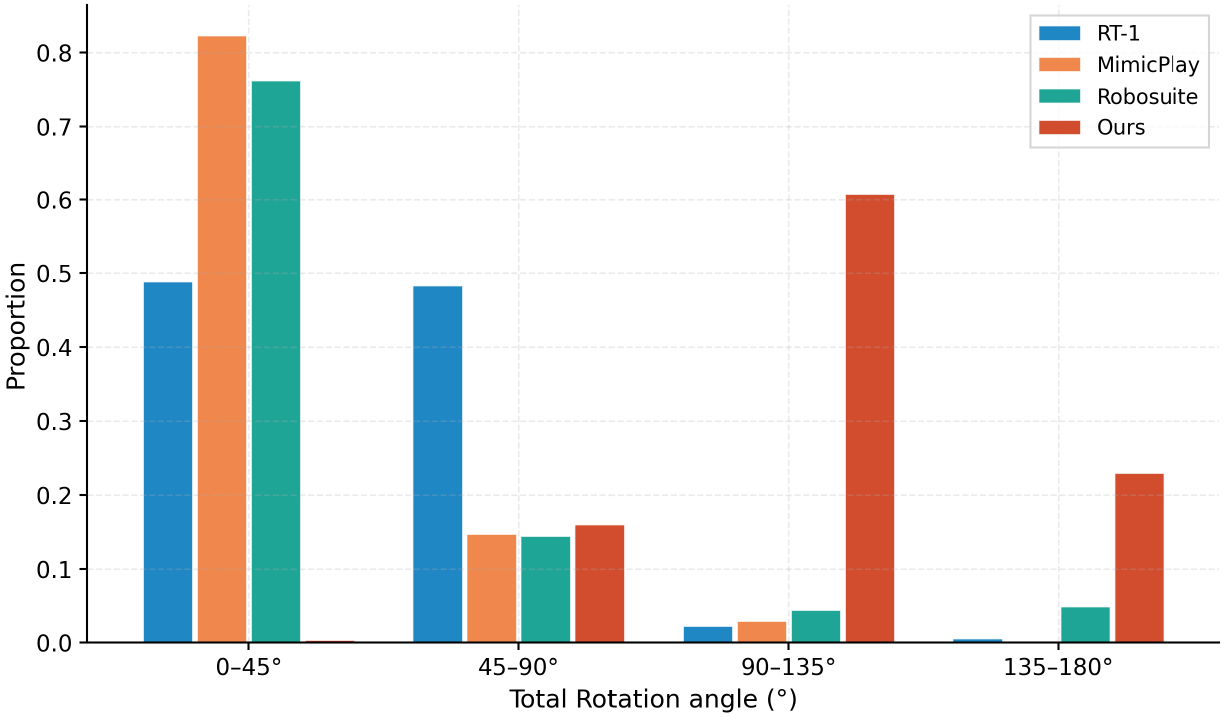}
  \caption{Histogram of Orientation Ranges Across Datasets}
  \label{fig:statis}
\end{figure}

While effective robot learning relies on large-scale, high-quality action data, acquiring such datasets remains time-consuming and limited in task diversity. Recently, leveraging human demonstrations—which inherently capture essential manipulation dynamics—has emerged as a more efficient and cost-effective alternative~\cite{zhao2025tasterobadvancingvideogeneration,wang2023mimicplaylonghorizonimitationlearning,bharadhwaj2024gen2acthumanvideogeneration,dessalene2025embodiswapzeroshotrobotimitation,chen2025toolasinterfacelearningrobotpolicies,park2025demodiffusiononeshothumanimitation,tang2025trajectoryconditionedcrossembodimentskill}. 
However, due to fundamental disparities in morphology, kinematics, and embodiment, existing works are largely constrained to relatively simple or planar object manipulations (\eg, pick-and-place, pushing). They often fail to facilitate \textit{complex spatial movements} (\eg, intricate trajectories with rotation or flipping), which are ubiquitous in everyday human demonstrations and essential for executing complex robotic tasks that require significant 3D traversal and orientation adjustments (as shown in Fig.~\ref{fig:statis}).

\begin{figure}[htbp]
  \centering             
  \includegraphics[width=\linewidth, keepaspectratio]{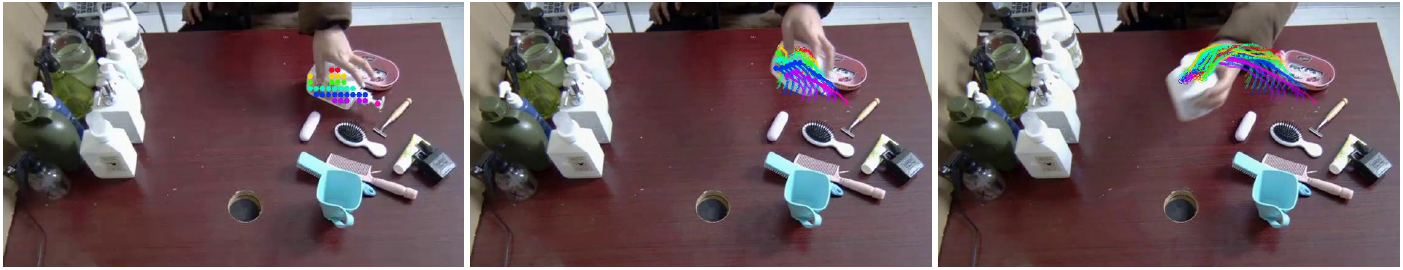}
  \vspace{-8pt}
  \caption{\textbf{Point tracking fundamentally fails under complex spatial movements, which severely occlude tracked points.} When tracking objects in manipulation videos, large rotations occlude tracked side points, forcing the tracker to drift onto the bottleneck. This occlusion-induced drift constitutes tracking failure, severely degrading the resulting flows.}
  \label{fig:track}
  \vspace{-10pt}
\end{figure}

\begin{figure}[htbp]
  \centering             
  \includegraphics[width=0.9\linewidth, keepaspectratio]{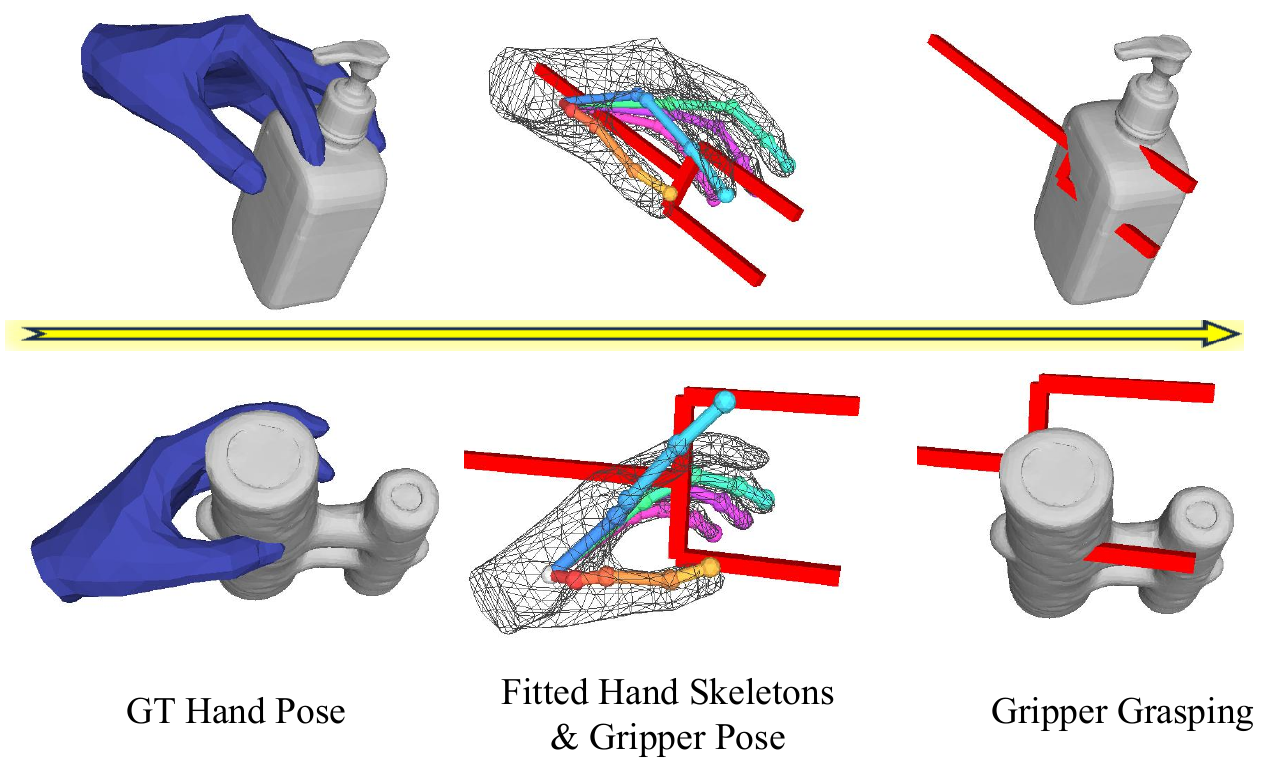}
  \caption{Failed results of rule-based retargeting methods when applied to natural hand grasping.}
  \label{fig:phantom}
\end{figure}

Specifically, existing approaches generally follow two paradigms. 
The first involves rule-based retargeting~\cite{lepert2025phantomtrainingrobotsrobots,zhou2025teachoncelearnoneshot,dessalene2025embodiswapzeroshotrobotimitation}, which maps keypoints from human hands to robotic grippers based on \textit{hand-crafted} kinematic heuristics. 
Nevertheless, as morphological disparities become particularly acute during complex spatial movements (\eg, Fig.~\ref{fig:phantom}), these hand-crafted mappings often fail to bridge the significant embodiment gap.
Secondly, \textit{trajectory-conditioned} (optimization-based~\cite{zhi20253dflowactionlearningcrossembodimentmanipulation,tang2025mimicfuncimitatingtoolmanipulation} and learning-based~\cite{xu2024flowcrossdomainmanipulationinterface,bharadhwaj2024track2actpredictingpointtracks}) methods attempt to guide robots using embodiment-agnostic (\ie, object-centric) motion paths extracted from human demonstrations. Nonetheless, it is difficult to reliably extract these trajectories in cluttered environments.
More importantly, they fail to encode the fine-grained dynamics inherent in human hand poses, which are critical for complex manipulation. Such hand-pose guidance is essential for tasks requiring precise spatial movement—for instance, re-orienting a gripper to avoid collisions or rotating a tool to maintain functional contact (\eg, Fig.~\ref{fig:track}).

To overcome the limitations of these existing paradigms and enable effective hand-to-gripper transfer for complex spatial manipulations, we raise two key insights: \textit{i) From hand-crafted to \textbf{data-driven} mapping:} Instead of relying on rigid, explicit kinematic rules, we shift toward an implicit, data-driven mapping. This approach allows the model to inherently learn the complex correspondences between human hand and robot gripper from \textit{large-scale data}.
\textit{ii) From trajectory-only to \textbf{hand-pose guidance}:} Recognizing that gripper-only or object-centric trajectories are insufficient for complex tasks, we argue that human hand-pose guidance is indispensable, advocating the use of hand-gripper \textit{paired data} to facilitate the learning of fine-grained pose dynamics.


Based on these two insights, we propose a scalable paired hand-gripper data acquisition pipeline, focusing on complex spatial movement. 
At the hand level, we design a rigorous protocol that prioritizes the diversity and complexity of spatial movements. Specifically, each manipulation episode incorporates varied trajectory and orientation transformations—such as horizontal rotations and vertical flips—guided by spatial markers within the operational workspace to ensure comprehensive complexity. Furthermore, for each object, we enhance motion diversity by utilizing different grasp gestures on different functional areas.
At the gripper level, we employ a handheld gripper (\eg, UMI \cite{chi2024universalmanipulationinterfaceinthewild}) for data collection. This interface is easy to control and operate, allowing the gripper to seamlessly mimic complex human hand motions. 
To efficiently acquire action annotations, we utilize a depth camera to capture monocular RGB-D videos. We then extract high-fidelity 3D action data—including 6-DoF gripper poses, hand and object points with contact maps registered in a unified global frame—by leveraging state-of-the-art pose estimation and tracking models~\cite{wen2024foundationposeunified6dpose,foundationposeplusplus}. This effectively bypasses the need for frequent recalibration when transitioning between different interaction environments.
Utilizing this pipeline, we collected a dataset including 6,189 episodes across 1,254 unique objects. As summarized in Sec.~\ref{sec:statistics}, our dataset exhibits significantly more complex spatial movements compared to existing hand-to-robot transfer benchmarks.

Building upon this dataset, we explore a data-driven approach to learn hand-to-gripper transfer. Our preliminary experiments indicate that brute-force end-to-end learning—directly mapping object and hand points to gripper actions—fails to achieve the precision required for reliable grasping, as evidenced in Tab.~\ref{tab:stage-ablation}. 
As a practical design to address this, we adopt a two-stage framework (see Fig.~\ref{fig:overview}): 
Stage I simplifies the learning objective by generating gripper poses specifically for the starting and terminal frames; Stage II then produces the full continuous action sequence, conditioned on these sparse frame-level generations. Furthermore, to mitigate cumulative drift in the generated gripper sequences, we disentangle the Stage II generation through a decoupled strategy: First, the gripper's central translational trajectory is derived from hand manipulation sequences and optimized based on kinematic consistency and the heuristic that grasping regions should be aligned for both hand and grippers; Conversely, the gripper’s orientation is learned implicitly from the paired data, as it lacks a simple geometric heuristic and represents the fundamental complexity of the hand-to-gripper transfer.

Leveraging this two-stage design, our framework infers directly from monocular hand RGB-D manipulation videos, where depth is obtained from either real-world capture or model prediction~\cite{chen2025videodepthanythingconsistent}. In both \textit{simulation} and \textit{real-robot} experiments, we conduct comprehensive comparisons against optimization-based and learning-based methods, and observe consistent improvements on tasks involving complex spatial movements.

Our contributions are summarized as:
\begin{itemize}
[itemsep=2pt,topsep=2pt,parsep=0pt, leftmargin=2em]
    \item To the best of our knowledge, the first hand-gripper dataset focusing on complex spatial movements.
    \item A scalable paired hand-gripper data acquisition pipeline that prioritizes manipulation complexity, with a rigorous UMI-based collection protocol and efficient annotation scheme. 
    \item A two-stage data-driven framework that learns effective hand-to-gripper transfer for complex manipulation movements. We view this as a 
    practical design enabled by our paired data.
    \item Our dataset, pipeline and algorithm will all be public.
\end{itemize}


\section{Related Works}

\subsection{Cross-embodiment Learning from Humans}
Learning from human demonstrations offers a cost-effective and scalable alternative to expensive robot-collected data. While some recent works have explored leveraging human-centric video demonstrations to train robot policy~\cite{haldar2025pointpolicyunifyingobservations,ren2025motiontracksunifiedrepresentation,lepert2025masqueradelearninginthewildhuman,lepert2025phantomtrainingrobotsrobots}, others aim to enable robots to perform a novel task with guidance from a single human demonstration, denoted as one-shot imitation~\cite{wang2023mimicplaylonghorizonimitationlearning,bharadhwaj2024gen2acthumanvideogeneration,dessalene2025embodiswapzeroshotrobotimitation,chen2025toolasinterfacelearningrobotpolicies,park2025demodiffusiononeshothumanimitation,tang2025trajectoryconditionedcrossembodimentskill}. 
Among these one-shot imitation methods, a common approach~\cite{lepert2025phantomtrainingrobotsrobots,zhou2025teachoncelearnoneshot,dessalene2025embodiswapzeroshotrobotimitation} involves kinematically retargeting human hand poses to robot end-effector poses at each timestep. While straightforward to implement, this hand-crafted method suffers from errors induced by the human-robot embodiment disparity. 
Another common approach - trajectory-conditioned policies
~\cite{xu2024flowcrossdomainmanipulationinterface, zhi20253dflowactionlearningcrossembodimentmanipulation,yuan2024generalflowfoundationaffordance} offer embodiment-agnostic, object-centric representations, they remain limited either by the inability of 2D trajectories~\cite{xu2024flowcrossdomainmanipulationinterface} to perceive orientation changes, by the fact that 3D trajectories~\cite{zhi20253dflowactionlearningcrossembodimentmanipulation} still recover poses via SVD over flow-tracked points, which degrades under drift in complex spatial movements. 
Therefore, a promising alternative is to adopt a data-driven approach.
While Human2Robot~\cite{xie2025human2robotlearningrobotactions} builds paired datasets, the robot data are collected via cumbersome teleoperation, limiting the demonstrations to simple tasks in clean environments with minimal spatial movements. This severely constrains scalability, and thus models trained on this data cannot generalize to casual human demonstrations in the wild. To overcome these limitations, we construct a scalable, high-quality human-robot paired data via UMI~\cite{chi2024universalmanipulationinterfaceinthewild}, focusing on complex spatial movements in object manipulations.

\subsection{Data Collection for Robot Learning}

A conventional method for collecting robot demonstrations is teleoperation~\cite{mandlekar2018roboturkcrowdsourcingplatformrobotic,zhao2023learningfinegrainedbimanualmanipulation,wu2024gellogenerallowcostintuitive,fu2024mobilealohalearningbimanual}, where a human operator directly controls the robot to generate task data. A growing body of VR-based research prioritizes efficient robot~\cite{iyer2024openteachversatileteleoperation,ding2024bunnyvisionprorealtimebimanualdexterous}, while another direction leverages VR primarily for acquiring paired human-robot demonstrations to support imitation learning~\cite{xie2025human2robotlearningrobotactions}. However, teleoperation is inherently labor‑intensive, difficult to execute for intricate tasks. Recently, the development of the UMI~\cite{chi2024universalmanipulationinterfaceinthewild} — a hand-held data collection device has enabled convenient and scalable acquisition of robot data. Several studies have enhanced UMI by integrating additional sensors for richer multimodal observations, such as tactile sensors~\cite{zhu2025touchwildlearningfinegrained} and depth sensors for point cloud capture~\cite{zhaxizhuoma2025fastumiscalablehardwareindependentuniversal,huang2025umigenunifiedframeworkegocentric}. We employ UMI to facilitate the scalable acquisition of seamlessly paired human-robot datasets focusing on complex object movements. Moreover, a depth camera is used to extract 3D action data, bypassing the need for frequent recalibration when changing interaction environments.

\subsection{Hand-Conditioned Grasping Detection and Generation}

Recent research on task‑oriented grasp generation for parallel‑jaw grippers has increasingly focused on learning directly from human hand demonstrations. Early end‑to‑end mapping methods~\cite{lepert2025phantomtrainingrobotsrobots,zhou2025teachoncelearnoneshot,dessalene2025embodiswapzeroshotrobotimitation,yuan2025demograspuniversaldexterousgrasping} often rely on hand‑crafted rules. For instance, Phantom~\cite{lepert2025phantomtrainingrobotsrobots} assume a fixed mapping — such as using the midpoint between thumb and index finger as the grasp point — which fails to handle the diversity of human grasps and may yield unstable robotic grasps. In contrast, later learning-based approaches~\cite{dong2024rtagrasplearningtaskorientedgrasping,ju2024roboabcaffordancegeneralizationcategories,Heppert_2024,cai2024visualimitationlearningtaskoriented} decouple grasp generation into two stages: they first sample task‑agnostic candidates~\cite{sundermeyer2021contactgraspnetefficient6dofgrasp,fang2020graspnet} and then filter them using human‑derived constraints, such as region~\cite{ju2024roboabcaffordancegeneralizationcategories,tang2025functofunctioncentriconeshotimitation} or combined region‑orientation constraints~\cite{dong2024rtagrasplearningtaskorientedgrasping}. While effective, this pipeline requires extensive sampling to satisfy both stability and task requirements, limiting efficiency. More recent work explores one‑stage generative models~\cite{huang2025hgdiffuserefficienttaskorientedgrasp,shi2025hograspflowexploringvisionbasedgenerative} that directly generate grasp poses conditioned on hand observations, unifying generation and task alignment. We extend this insight to action sequence generation, as task instruction is often expressed through sequences of hand manipulation.

\section{Dataset}
\label{sec:dataset}


In our dataset, (1) we adopt a highly scalable collection strategy, including using handheld grippers UMI~\cite{chi2024universalmanipulationinterfaceinthewild} and only recording paired RGB-D videos (Sec~\ref{sec:collection}) from a fixed viewpoint. (2) we extract 3D motion data from raw RGB-D videos for cross-embodiment learning (Sec~\ref{sec:extraction}). (3) In comparison with other datasets, the spatial movement of our implemented tasks is more diverse and complex (Sec~\ref{sec:statistics}).

\subsection{Data Collection}
\label{sec:collection}

\begin{figure*}[!htbp]
  \centering             
  \includegraphics[width=0.8\linewidth, keepaspectratio]{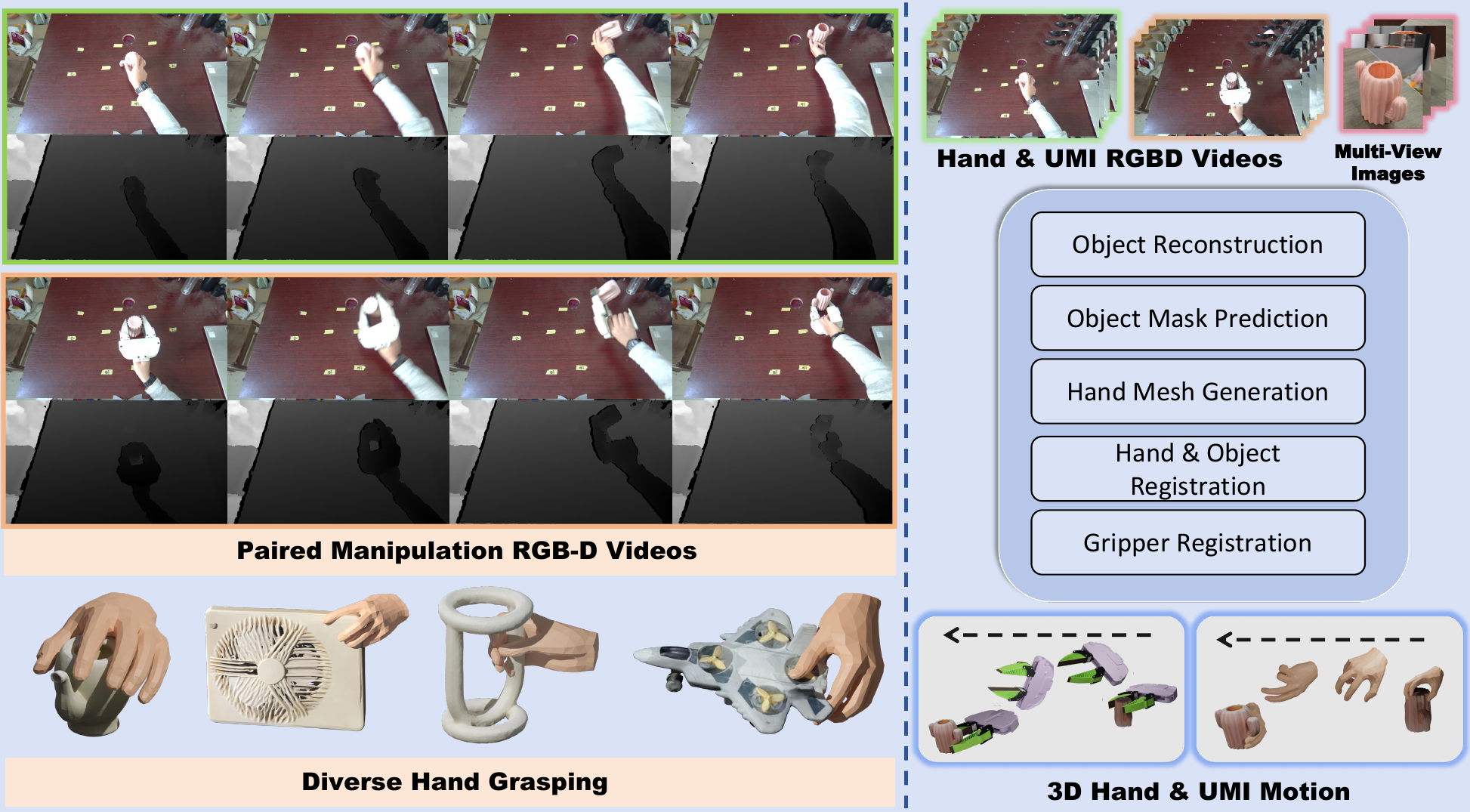}
  \caption{\textbf{Dataset collection and processing.} \textbf{(L)} We record paired RGB-D videos of human hand and gripper manipulation under diverse grasping poses, providing paired data for hand-to-gripper learning. \textbf{(R)} From RGB-D videos and multi-view object captures, we extract registered 3D motion data — including object point clouds, hand meshes, and gripper actions — sufficient for hand-to-gripper learning.}
  \label{fig:dataset}
\end{figure*}

The enhanced diversity of our dataset manifests in two key aspects, \textbf{natural grasping types in human videos}: by encompassing a wide spectrum of natural hand grasping types (not only restricted to pinch), we significantly enhance the model's robustness for generalizing across in-the-wild human demonstrations; and \textbf{diverse orientation transformation}: the inclusion of complex in-hand rotations enables the learning of sophisticated manipulation skills beyond simple translation. To achieve the above points, we adopt the following strategy: (1) As shown in Fig.~\ref{fig:dataset} (L), when hand manipulating, collectors actively vary their hand gestures and contact areas on the objects without any constraints. (2) As shown in Fig.~\ref{fig:dataset} (L), each task of our dataset incorporates complex orientation transformations — including in-plane rotation and vertical flipping — designed to achieve precise placement in cluttered environments. (3) To collect paired gripper videos efficiently, collectors use a hand-held gripper UMI~\cite{chi2024universalmanipulationinterfaceinthewild} to complete the identical task. Specifically, the UMI manipulation aims to closely imitate the contact area, grasping orientation, and manipulation motions observed in the human demonstration video. 
(4) After recording, we verify that the object poses at the starting and terminal frames are consistent between human and UMI demonstrations. Trials with significant misalignment are filtered out, and only consistent pairs are retained.
Using this strategy, we collected a dataset comprising \textbf{6,189} manipulation episodes across \textbf{1,254} distinct object instances.

\subsection{Extraction of 3D Motion Data}
\label{sec:extraction}

For scalability, we directly extract 3D motion data from recorded RGB-D videos instead of complex calibration-based acquisition methods. Our extraction pipeline follows four steps: (1) Object mesh reconstruction from multi-view images using ReconViaGen~\cite{chang2025reconviagenaccuratemultiview3d}, (2) Registering objects in 3D scenes, (3) Reconstructing and registering hand meshes in 3D scenes, and (4) Registering and tracking UMI in 3D scenes. During the entire process, we employ FoundationPose++~\cite{foundationposeplusplus} to register and track using RGB-D frames, corresponding mesh and an initial mask image as input. Finally, as shown in Fig.~\ref{fig:dataset} (R), we obtain a 3D motion dataset comprising object point clouds, hand mesh sequences, and the UMI 6-DOF pose sequence, which is sufficient for hand-to-gripper learning. We also present some visualization in Appendix~\ref{sec:register-vis}.

\paragraph{\textbf{Objects and hands.}} Given that the first and last frames are free of occlusion from hands and UMI, we leverage their frames to perform registration. Specifically, we utilize a hand-object detector~\cite{shan2020understandinghumanhandscontact} to detect the target object's bounding box, which is then used to predict its mask image via SAM2~\cite{ravi2024sam2segmentimages}. For the hands, due to the slight variations in hand pose during manipulation, we forgo tracking and instead utilize Wilor~\cite{potamias2025wilorendtoend3dhand} to reconstruct hand meshes and masks for each sampled frame, which are then directly registered in the 3D scene.

\paragraph{\textbf{Registering  and tracking UMI}} For UMI manipulation videos, we first annotate the \textit{starting} frame (when UMI initially grasps the object) and the \textit{terminal} frame (when UMI finishes manipulation and places the object). Next, we predict a mask image of the starting frame using SAM2~\cite{ravi2024sam2segmentimages} via click-based interaction. Finally, we register and track the full manipulation sequence with predicted mask and UMI CAD mesh, yielding 6-DOF pose sequence that directly serves as robot action data. 

\paragraph{Dataset Pairing.}
To ensure the spatial alignment, we compute a 3D trajectory similarity score (details in Appendix.~\ref{sec:TS}) for each hand-UMI pair, measuring the spatial alignment between the hand and UMI trajectories, and only retain pairs whose similarity exceeds 0.9. This procedure yields high-quality pairs with a mean similarity of 0.957 (median 0.958), indicating that the retained pairs are closely matched. The similarity distribution of retained pairs is provided in the appendix (Fig.~\ref{fig:TS}).

\subsection{Dataset Statistics}
\label{sec:statistics}

In our dataset, we collect 3 to 5 paired manipulation episodes for each object. In total, our dataset comprises \textbf{6,189} episodes across \textbf{1,254} unique objects. Our dataset is characterized by Complex Spatial Movements, quantified by metrics such as the total rotation angle during manipulation. This stands in contrast to prevalent real-world robotic datasets (e.g. Robosuite~\cite{zhu2020robosuitemodularsimulationframework}, MimicPlay~\cite{wang2023mimicplaylonghorizonimitationlearning} and RT-1~\cite{brohan2023rt1roboticstransformerrealworld}), which predominantly consist of primarily translational actions like pushing, sliding, or pick-and-place with minimal reorientation. As shown in Fig.~\ref{fig:statis}, the orientation variation in our dataset substantially exceeds that of prior datasets.

\begin{table}[H]
\caption{Distribution of Object Categories in Data Collection}
\label{tab:object}
\begin{minipage}{\columnwidth}
\begin{center}
\begin{tabular}{lllll}
  \toprule
  Food & Decor & Beauty & Necessities & Toys \\ \midrule
  23.03\%                & 27.88\% & 3.64\% & 12.12\% & 33.33\%\\
  \bottomrule
\end{tabular}
\end{center}
\bigskip\centering
\end{minipage}
\vspace{-25pt}
\end{table}

To enhance the model generalization on unseen objects, we collected a diverse and comprehensive set of objects (as shown in Tab.~\ref{tab:object}). Specifically, the five categories and their corresponding object types are as follows: F (Food) covers edible items such as egg tarts, fried dough sticks, milk toast, and ice cream, involving both baked goods and snacks; D (Decor) includes various decorative ornaments like bronze gun figurines, nine-petal flower petal ornaments, golden horse figurines, and shell conch decorations; B (Beauty) consists of beauty and personal care products such as light purple makeup brushes and matte red lipsticks; N (Necessity) contains daily necessities and storage tools, for example, three-layer blue drawers, solid wood black watch stands, and ocean blue spray bottles; T (Toy) involves a wide range of model toys, including white camera models, green train models, yellow excavator models, and Plants vs. Zombies zombie figurines.

\section{Method}
\label{sec:method}

\begin{figure*}[htbp]
  \centering             
  \includegraphics[width=0.7\linewidth, keepaspectratio]{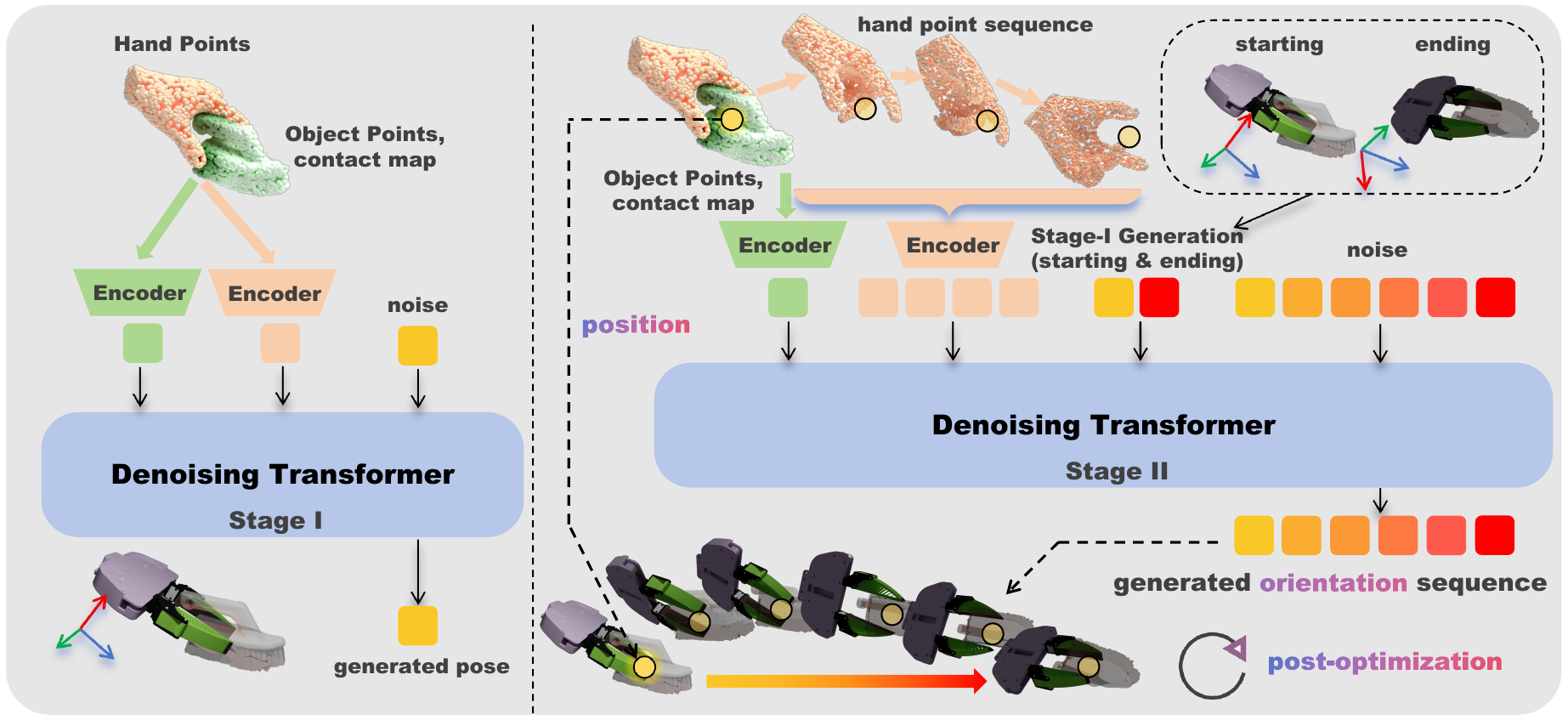}
  \caption{Overview of the framework. CosmoH2G consists of a two-stage framework: (L) the model generates single-frame action conditioned on given HOI point clouds. During inference, the model generates poses for the starting and terminal frames, respectively, (R) the model generates orientation sequence conditioned on object points, hand points sequence and Stage-I generation. During inference, the pose sequence is produced by combining the generated orientation sequence with the position sequence calculated from the hand sequence, under the prior that the grasping regions (shown as the small yellow circle) are approximately aligned. Additionally, we further optimize the sequence by considering contact-map alignment, trajectory smoothness, and physical plausibility.}
  \label{fig:overview}
\end{figure*}

We aim to transfer hand motions in human videos to robot actions, which are always end-effector 6-DOF pose sequences. 
Both previous optimization-based methods~\cite{lepert2025phantomtrainingrobotsrobots,zhou2025teachoncelearnoneshot,dessalene2025embodiswapzeroshotrobotimitation,zhi20253dflowactionlearningcrossembodimentmanipulation} and trajectory-based methods~\cite{xu2024flowcrossdomainmanipulationinterface,bharadhwaj2024track2actpredictingpointtracks} exhibit degraded performance under complex spatial movements. 
Therefore, we employ a purely data-driven approach to learn the transfer function from our paired dataset without relying on any pre-defined alignments. To enhance performance, we adopt a two-stage framework: (1) Stage I predicts the starting and terminal 6-DOF gripper poses (Sec.~\ref{sec:stageI}); (2) Stage II generates the pose sequence for the entire manipulation (Sec.~\ref{sec:stageII}). An overview of our framework is illustrated in Fig~\ref{fig:overview}. 

\subsection{Problem Formulation}
\label{sec:formulation}

Given a single RGB-D video of human demonstration (depth from either real capture or model prediction), we focus on enabling robots to learn and execute the demonstrated task. Specifically, a human performs a manipulation task, recording a sequence of RGB-D frames, $V^h = \{I_i^h\}_{i=0}^{N^h-1}$, where $N^h$ denotes the total frames. Through data processing (as illustrated in sec~\ref{sec:extraction}), we can obtain corresponding 3D data, including: (1) object points at starting frames $p^o_{i=0}$ and terminal frames $p^o_{i=-1}$, and (2) hand points sequence $\{p_i^h\}_{i=0}^{N^h-1}$ sampled from hand MANO~\cite{romero2022embodied} meshes. 
In Stage I, the model independently generates single-frame 6-DOF grasping poses for the starting frame ($g_{i=0} \in SE(3)$) and the terminal frame ($g_{i=-1} \in SE(3)$), with each conditioned on the hand points, object points, and their corresponding contact map (details in Appendix~\ref{sec:contactmap}). 
In Stage II, the model aims to generate gripper orientation sequence, conditioned on hand points sequence, object points, their corresponding contact map, and the Stage-I generation. The gripper position sequence is then computed from hand points sequence under the prior that grasping regions are approximately aligned, and subsequently optimized.


\subsection{Stage I: Generating Starting and Terminal Action}
\label{sec:stageI}

In Stage I, we train a denoising Transformer~\cite{peebles2023scalablediffusionmodelstransformers} to independently generate single-frame 6-DOF gripper pose for the starting and terminal frames, conditioned on the hand points, object points, and their corresponding contact map. As shown in Fig~\ref{fig:overview}(L), for better modeling, we separately encode the hand points and object points into their respective feature vectors, concatenate them along with the noisy action, and represent each as a single token. These tokens are then processed by a denoising transformer to generate the clean action through multi-step denoising.

\paragraph{\textbf{Point Features.}} We employ two separate Point Transformer encoders~\cite{zhao2021pointtransformer}, parameterized by $\theta_1$ and $\theta_2$, to extract global features from the object and hand point clouds, respectively. 
Specifically, the object encoder $\mathcal{E}^o_{\theta_1}$ takes the object points $p^o \in \mathbb{R}^{M_o \times 3}$ along with their associated contact map $m \in \mathbb{R}^{M_o}$ appended as an optional per-point feature channel, producing the object feature $f^o = \mathcal{E}^o_{\theta_1}(p^o, m) \in \mathbb{R}^{d}$. 
The hand encoder $\mathcal{E}^h_{\theta_2}$ processes only the hand points $p^h \in \mathbb{R}^{M_h \times 3}$, yielding the hand feature $f^h = \mathcal{E}^h_{\theta_2}(p^h) \in \mathbb{R}^{d}$. 
Here, $M_o$ and $M_h$ denote the number of points in the object and hand point clouds, respectively, and $d$ is the feature dimension. Notably, our method only requires a coarse contact area rather than precise finger-level contact, for two reasons. First, contact errors from off-the-shelf hand estimators (e.g., WiLoR) are mostly localized to finger regions, whereas our method merely needs to know which object regions are grasped. Second, given the structural gap between a five-fingered hand and a two-finger gripper, fine-grained finger placements are irrelevant to the transfer.

\paragraph{\textbf{Action Representation.}} Previous work typically represents robot actions using either a Homogeneous Transformation Matrix or a Quaternion-Translation pair. 
Inspired by HGDiffuser~\cite{huang2025hgdiffuserefficienttaskorientedgrasp}, we use the three keypoints of gripper (the gripper center and the fingertips of the left and right fingers) to parameterize its SE(3)  pose (see visualization in Appendix~\ref{sec:keypoint-vis}). 
A key advantage is that conditioning and generation share the same 3D-point space, eliminating cross-domain mapping and simplifying learning.
To enable the action to be processed alongside other conditioning feature tokens, we project the keypoints' coordinates into the feature space using a linear layer. After denoising, a separate linear layer maps the feature back to the action space.

We train the model, denoted as $\mathcal{G}_{\eta_1}$,  by
optimizing a mean squared error objective:

\begin{equation}
L_{MSE} = \mathbb{E}_{a_0,t}\left[\|a_0 - \mathcal{G}_{\eta_1}(a_t, t, f^o, f^h)\|^2_2\right],
\label{eq:stage1-loss}
\end{equation}
where $a_0 \in \mathbb{R}^{3\times3}$ denotes the ground-truth (clean) 
action in the three-keypoint representation, $a_t$ is its noisy version 
at diffusion timestep $t$ with $T=1000$, 
and $f^o, f^h$ are the object and hand point-cloud features.


\subsection{Stage II: Predicting Action Sequences}
\label{sec:stageII}

In Stage II, we aim to generate the full gripper action sequence. A straightforward approach would be to train a denoising Transformer to directly generate the pose sequence, encompassing both position and orientation. 
To mitigate cumulative drift in the generated gripper sequences, we adopt a factorized formulation: the model is trained to learn solely the orientation mapping, while the position sequence is derived from hand manipulation videos and further refined via optimization.

\paragraph{\textbf{Generation of Orientation Sequence.}} As illustrated in Sec.~\ref{sec:stageI}, we use the absolute coordinates of keypoints to represent robot actions. Specifically, the gripper position is given by the center keypoint, while the orientation is derived from the relative positions of the left and right finger keypoints with respect to this center. 
Accordingly, in Stage II, the model learns only the orientation mapping. The ground-truth supervision is obtained by computing these relative coordinates directly from the gripper keypoints, and the model is trained to generate the gripper orientation sequence in the same representation. To this end, the model conditions on the object points and contact map, the hand points sequence, and the starting and terminal poses generated by Stage I. As shown in Fig~\ref{fig:overview} (R), 
each frame of hand points sequence and each frame of the noisy orientation sequence is encoded into its own token. Likewise, the starting and terminal poses are each treated as distinct tokens. All these tokens are fed into the Transformer, which progressively denoises to generate the clean orientation sequence.


We train the model, denoted as $\mathcal{G}_{\eta_2}$,  by
optimizing a mean squared error objective:

\begin{equation}
L_{MSE} = \mathbb{E}_{s_0,t}\left[\|s_0 - \mathcal{G}_{\eta_2}(s_t, t, f^o, 
\{f^h_i\}_{i=1}^{N_h}, \hat{a}_0, \hat{a}_{-1})\|^2\right],
\label{eq:stage2-loss}
\end{equation}
where $s_0$ denotes the ground-truth orientation sequence in the 
three-keypoint representation, $s_t$ is its noisy version at diffusion 
timestep $t$, and $\hat{a}_0, \hat{a}_{-1}$ are the starting and 
terminal actions generated by Stage I.


\paragraph{\textbf{Optimization of Position Sequence.}}
During inference, we initialize the gripper position sequence from the hand trajectory under the prior that their contact regions are spatially aligned.
This yields a physically unconstrained initialization, which we refine through a three-step optimization. First, we optimize the gripper positions at starting and terminal frames to ensure stable grasping. The objective is formulated as:

\begin{equation}
\Delta x^* = \arg\min_{\Delta x}\ \Big[L_{contact} + \lambda_{pen} L_{pen} 
+ \lambda_{reg} \|\Delta x\|^2\Big],
\label{eq:keyframe-opt}
\end{equation}
where $\Delta x$ denotes the correction applied to the gripper 
positions at the starting and terminal frames, $L_{contact}$ encourages 
the gripper to approach the hand contact region, $L_{pen}$ penalizes 
object penetration, and $\lambda_{pen}, \lambda_{reg}$ are weighting 
coefficients. Second, with the optimized starting and terminal positions fixed, we refine the intermediate frames to ensure a smooth trajectory while staying close to the initialization:


\begin{equation}
\{x_i^*\} = \arg\min_{\{x_i\}}\ \lambda_{smooth} L_{smooth} 
+ \lambda_{ref} L_{ref},
\label{eq:traj-opt}
\end{equation}
where $L_{smooth} = \sum_i \|x_{i-1} - 2x_i + x_{i+1}\|^2$ penalizes 
abrupt motions, and $L_{ref} = \sum_i \min_{s\in[0,1]} \|x_i - 
C_{ref}(s)\|^2$ pulls each frame toward a reference curve 
$C_{ref}: [0,1] \to \mathbb{R}^3$ interpolated from the initialized 
positions, with $s \in [0,1]$ the normalized curve parameter and 
$\min_s$ selecting the closest point on the curve.
Third, to ensure kinematic feasibility, we optimize each gripper pose via an IK solver. The objective minimizes the end-effector residual and joint-limit violations while regularizing the pose to stay close to the initialization:

\begin{equation}
a_i^* = \arg\min_{a_i}\ \Big[L_{ik}(a_i) + \lambda_{reg}\|a_i - \bar{a}_i\|^2\Big],
\label{eq:ik-opt}
\end{equation}
where $\bar{a}_i$ is the initialized pose at frame $i$, composed of the 
optimized position $x_i^*$ (Eq.~\ref{eq:traj-opt}) and the generated 
orientation, $L_{ik}$ penalizes the error between the IK solution and 
the target pose as well as joint-limit violations, and the second term 
regularizes the pose to stay close to the initialization.


\subsection{Implementation Details}
\label{sec:implementation}

Both stages employ a Transformer-based diffusion model~\cite{peebles2023scalablediffusionmodelstransformers} with separate Point Transformer encoders~\cite{zhao2021pointtransformer} for object and hand point clouds, producing $256$-dimensional features. The gripper pose is parameterized by three keypoints and projected into token space via linear layers. We train both denoising Transformers from scratch using AdamW with a fixed learning rate of $10^{-4}$, batch size $32$ on a single NVIDIA A100 for $20{,}000$ steps. The diffusion process uses 1000 timesteps with a cosine schedule; during inference, we apply DDPM sampling with 500 steps. 

\section{Evaluation}

We design comprehensive experiments to evaluate our
proposed \textbf{CosmoH2G} in terms of (1) Oriented grasping guided by human demonstration, (2) Trajectory adherence and final placement accuracy, (3) Target orientation placement accuracy.

\begin{table*}[!htbp]
\caption{Quantitative Comparison to Baselines in Simulation and Real-World Experiments.}
\label{tab:baseline}
\small
\begin{center}
\setlength{\tabcolsep}{4pt}
\begin{tabular}{l|cccc|cccc}
\toprule
& \multicolumn{4}{c|}{\textbf{Simulation}} & \multicolumn{4}{c}{\textbf{Real Robot}} \\
Methods & GOA$\downarrow$ & SR$\uparrow$ & TS$\uparrow$ & TOPA$\downarrow$ & GOA$\downarrow$ & SR$\uparrow$ & TS$\uparrow$ & TOPA$\downarrow$ \\
\midrule
\multicolumn{9}{c}{\cellcolor{gray!30}{ReTargeting-Optimization-based}} \\
\midrule
MimicFunc & 19.37$^\circ$ & 56.45\% & 0.7673 & 32.47$^\circ$ & 19.37$^\circ$ & 43.01\% & 0.7714 & 44.84$^\circ$ \\
3DFlowAction & 20.84$^\circ$ & 44.62\% & 0.6752 & 29.17$^\circ$ & 20.84$^\circ$ & 36.02\% & 0.7936 & 47.27$^\circ$ \\
\midrule
\multicolumn{9}{c}{\cellcolor{gray!30}{Learning-based}} \\
\midrule
Im2Flow2Act & 29.86$^\circ$ & 57.53\% & 0.7835 & 50.27$^\circ$ & 29.86$^\circ$ & 51.61\% & 0.7381 & 57.89$^\circ$ \\
Track2Act & 10.67$^\circ$ & 77.42\% & 0.9015 & 12.48$^\circ$ & 10.67$^\circ$ & 60.22\% & 0.8732 & 20.58$^\circ$ \\
\midrule
Ours & \textbf{7.53$^\circ$} & \textbf{83.87\%} & \textbf{0.9672} & \textbf{10.27$^\circ$} & \textbf{7.53$^\circ$} & \textbf{70.43\%} & \textbf{0.9413} & \textbf{19.34$^\circ$} \\
\bottomrule
\end{tabular}
\end{center}
\end{table*}

\subsection{Experimental Setups}

\paragraph{\textbf{Setup.}} To evaluate how well CosmoH2G generalizes to novel objects and environments, we split the dataset by object. The resulting test set comprises 186 cases spanning 40 objects that are unseen during training. 
Notably, we focus on pick-and-place tasks with unseen hand motions, leaving broader task types as future work.
We evaluate both in simulation and on real robots: In simulation, we evaluate on GalaxeaManipSim~\cite{GalaxeaManipSim} using R1 Lite embodiment. We reconstruct the full test-set episodes within the simulator, ensuring that object geometries, initial configurations, and target placements faithfully match the hand demonstrations in our dataset; On the real robot, we evaluate the same test-set tasks on the Galaxea R1 Lite (as shown in Fig.~\ref{fig:setup}). The complementary details are illustrated in Appendix.~\ref{sec:setup-detail}.


\paragraph{\textbf{Evaluation Metrics.}} The following metrics are evaluated for the aforementioned three aspects (more details in Appendix~\ref{sec:metric}):

\begin{enumerate}
\item \textbf{Oriented grasping guided by human demonstration.} We aim to assess robot's ability to replicate the demonstrated grasp orientation, as it reflects a feasible choice. The assessment is conducted by computing angular deviation between executed and ground truth grasping poses (\textbf{Grasping Orientation Accuracy (GOA)}). 

\item \textbf{Trajectory adherence and final placement accuracy.} Given the hand motion as a condition, we evaluate the robot's manipulation based on two metrics: \textbf{TS (Trajectory Similarity)} and \textbf{Success Rate (SR)}. This dual evaluation assesses both how closely the robot follows the demonstrated motion and how accurately it accomplishes the task.
\item \textbf{Target
orientation placement accuracy.} We quantify orientation accuracy via the angular error between the achieved and desired orientations of the manipulated object, with smaller values reflecting higher precision. This metric is named \textbf{Target Orientation Placement Accuracy (TOPA)}.
\end{enumerate}

\paragraph{\textbf{Baselines.}} Since the hand-pose retargeting methods are even unable to grasp objects under our tasks (shown in Fig.~\ref{fig:phantom}), we directly compare with trajectory-based methods: 
\begin{enumerate}
\item \textbf{Optimization-based methods}: These methods first calculate an action sequence through retargeting hand poses~\cite{lepert2025phantomtrainingrobotsrobots,zhou2025teachoncelearnoneshot,dessalene2025embodiswapzeroshotrobotimitation} or by directly applying the object's relative transformation~\cite{tang2025mimicfuncimitatingtoolmanipulation,zhi20253dflowactionlearningcrossembodimentmanipulation}, and then optimize it. 
Among these, we select MimicFunc~\cite{tang2025mimicfuncimitatingtoolmanipulation} and 3DFlowAction~\cite{zhi20253dflowactionlearningcrossembodimentmanipulation} as baselines. 

\item {\textbf{Learning-based methods}}: these methods always directly generate robot actions conditioned on tracked 2D trajectory flows of human demonstrations. Among these, we choose Im2Flow2Act~\cite{xu2024flowcrossdomainmanipulationinterface} and Track2Act~\cite{bharadhwaj2024track2actpredictingpointtracks} as baselines. Specifically, since Track2Act has not publicly released corresponding pretrained checkpoints, we re-train it on our dataset for comparison. 
\end{enumerate}

\subsection{Experimental Results}

\paragraph{\textbf{Quantitative Comparison to Baselines.}} Compared to baselines, our CosmoH2G  demonstrates superior performance in both simulation and real robot experiments. Real-robot performance is slightly lower, mainly due to two physical factors: (a) objects occasionally slip during large rotations, leading to orientation shifts or drops; (b) objects are sometimes moved before being fully lifted from the table, causing excessive friction and subsequent drops.

\begin{figure*}[htbp]
  \centering             
  \includegraphics[width=0.85\linewidth, keepaspectratio]{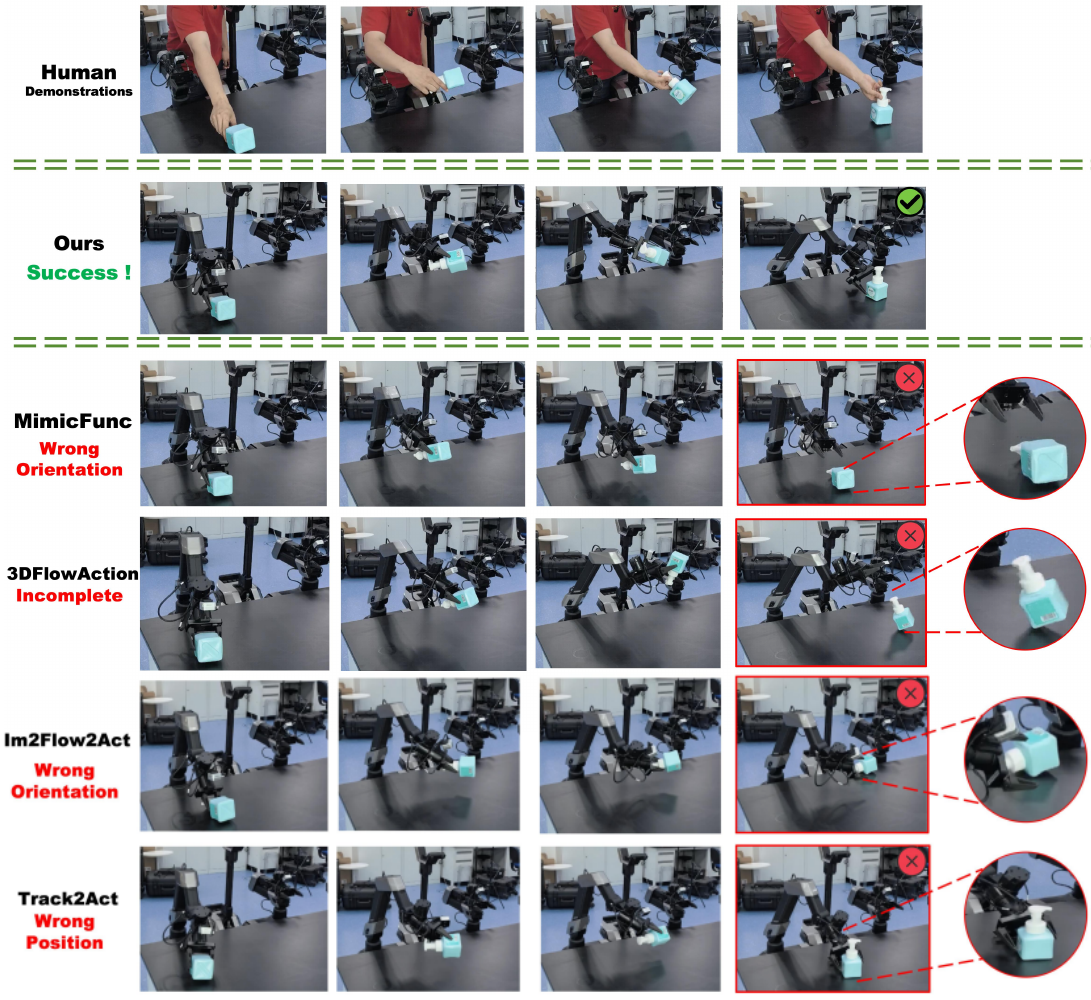}
  \caption{Quantitative Comparison with Baselines in real-robot experiments.}
  \label{fig:baselines-real}
\end{figure*}

For Mimicfunc~\cite{tang2025mimicfuncimitatingtoolmanipulation}, CoTracker's tracking degrades under complex spatial movements (Fig.~\ref{fig:track}), leading to incorrect action generation.
Similarly, 3DFlowAction~\cite{zhi20253dflowactionlearningcrossembodimentmanipulation} and im2flow2act~\cite{xu2024flowcrossdomainmanipulationinterface} rely on 2D tracking, which introduces errors that degrade their performance.
Despite fine-tuning Track2Act~\cite{bharadhwaj2024track2actpredictingpointtracks} on our dataset, its calculation-and-refinement framework leads to suboptimal predictions, particularly when the calculation deviates significantly from the desired actions — a common scenario in our tasks. We further note that none of the baselines takes a contact map as input, and our gains are not attributable to this additional information: even without the contact map, CosmoH2G still outperforms all baselines (Table~\ref{tab:contactmap}, ``w/o contact map''). These results confirm that the improvement stems from our formulation rather than the contact-map input itself, ensuring a fair comparison across baselines.


\subsection{Ablation Results}

\paragraph{\textbf{Single-Stage Versus Two-Stage.}} We conduct an ablation study to validate the necessity of our two-stage framework, as shown in Tab.~\ref{tab:stage-ablation}. For \textit{Single-Stage}, we adopt three distinct architectures: Diffusion Transformer~\cite{peebles2023scalablediffusionmodelstransformers}, Diffusion Policy~\cite{chi2023diffusionpolicy}, and ACT~\cite{zhao2023learningfinegrainedbimanualmanipulation}. All variants uniformly suffer
from insufficient accuracy, affirming that the performance degrada-
tion is a limitation of the evaluated single-stage formulation.


\begin{table}[H]
\caption{Ablation Study: \textit{Single-Stage} Versus \textit{Two-Stage} (in simulation).}
\label{tab:stage-ablation}
\begin{minipage}{\columnwidth}
\begin{center}
\begin{tabular}{lllll}
  \toprule
  Methods                     & GOA$\downarrow$   & SR$\uparrow$ & TS$\uparrow$ & TOPA$\downarrow$ \\ \midrule
\multicolumn{5}{c}{\cellcolor{gray!30}{Single-Stage}} \\
\midrule
  Denoising Transformer                & 15.28$^\circ$ & 62.90\% & 0.8542 & 47.83$^\circ$\\
  Diffusion Policy & 12.36$^\circ$ & 58.60\% & 0.8157 & 52.84$^\circ$\\
  ACT & 14.58$^\circ$ & 64.52\% & 0.8732 & 49.27$^\circ$\\
\midrule
  Ours (Two-Stage)                       & \textbf{7.53$^\circ$} & \textbf{83.87\%} & \textbf{0.9672} & \textbf{10.27$^\circ$}\\
  \bottomrule
\end{tabular}
\end{center}
\bigskip\centering
\end{minipage}
\end{table}

\paragraph{\textbf{Model Condition.}} We conduct an ablation study to validate the contribution of our designed condition, as shown in Tab.~\ref{tab:contactmap}. For \textit{contact map} (the condition of both stages), removing it leads to a slight performance drop, demonstrating that it provides valuable geometric cues for reliable and precise grasping. For \textit{ending action} (the condition of stage-II), while the starting action provides a kinematic prior for initialization, adding the ending action is essential for achieving the highest precision, as it constrains the terminal pose to ensure accurate placing orientation.


\begin{table}[H]
\caption{Ablation Study: Model Condition (in simulation).}
\label{tab:contactmap}
\begin{minipage}{\columnwidth}
\begin{center}
\begin{tabular}{lllll}
  \toprule
  Methods           & GOA$\downarrow$   & SR$\uparrow$ & TS$\uparrow$ & TOPA$\downarrow$ \\ \midrule
  w/o contact map   & 9.58$^\circ$ & 80.11\% & 0.9117 & 11.96$^\circ$\\
  w/o ending action & 9.48$^\circ$ & 76.34\% & 0.8932 & 21.79$^\circ$\\
  Ours              & \textbf{7.53$^\circ$} & \textbf{83.87\%} & \textbf{0.9672} & \textbf{10.27$^\circ$}\\
  \bottomrule
\end{tabular}
\end{center}
\bigskip\centering
\end{minipage}
\end{table}

\paragraph{\textbf{Representation of Robot Action.}} Different from natural format (either a Homogeneous Transformation Matrix
or a Quaternion-Translation pair), we use three keypoints of grippers to represent the robot action. We conduct an ablation study on it to evaluate how different action representations affect the transfer performance. As shown in Tab.~\ref{tab:representation}, the improvement stems from the explicit spatial nature of the keypoint representation, where the three points directly denote contact regions and orientation of the gripper. 

\begin{table}[H]
\caption{Ablation Study: Representation of robot action (in simulation).}
\label{tab:representation}
\begin{minipage}{\columnwidth}
\begin{center}
\begin{tabular}{lllll}
  \toprule
  Methods           & GOA$\downarrow$   & SR$\uparrow$ & TS$\uparrow$ & TOPA$\downarrow$ \\ \midrule
  Matrix   & 13.17$^\circ$ & 69.89\% & 0.8868 & 28.47$^\circ$\\
  Quat     & 11.34$^\circ$ & 72.58\% & 0.8984 & 24.63$^\circ$\\
  Ours(keypoints)              & \textbf{7.53$^\circ$} & \textbf{83.87\%} & \textbf{0.9672} & \textbf{10.27$^\circ$}\\
  \bottomrule
\end{tabular}
\end{center}
\bigskip\centering
\end{minipage}
\end{table}



\paragraph{\textbf{Post-optimization.}}
To validate the improvement achieved by \textit{post-optimization}, we compare three variants in Tab.~\ref{tab:ablation}. The \textit{Direct} baseline trains the model to generate the full pose sequence in an end-to-end manner. The \textit{Computed} variant derives both position and orientation purely from geometric heuristics and optimization. 
As shown, the \textit{Direct} baseline suffers from minor inaccuracies caused by imperfect hand-gripper trajectory alignment in the paired data.
Additionally, the \textit{Computed} variant yields inaccurate orientations, as the lack of a trustworthy geometric prior results in poor initialization, while the high-dimensional pose space further makes optimization highly challenging. 

\begin{figure*}[htbp]
  \centering             
  \includegraphics[width=0.7\linewidth, keepaspectratio]{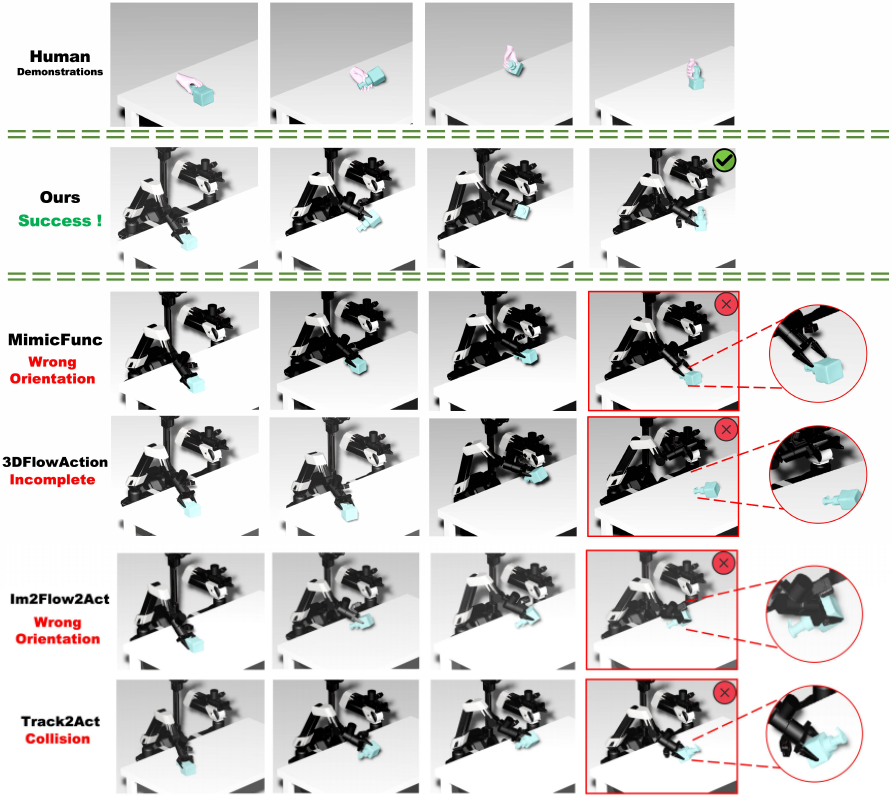}
  \caption{Quantitative Comparison with Baselines in simulation experiments.}
  \label{fig:baselines}
\end{figure*}

\begin{table}[H]
\caption{Ablation on Stage-II Generation Strategy (in simulation).}
\label{tab:ablation}
\small
\begin{center}
\setlength{\tabcolsep}{5pt}
\begin{tabular}{lcccc}
\toprule
\textbf{Variant} & GOA$\downarrow$ & SR$\uparrow$ & TS$\uparrow$ & TOPA$\downarrow$ \\
\midrule
Direct (pos + ori) & 8.50$^\circ$ & 81.72\% & 0.9176 & 13.23$^\circ$ \\
Computed (pos + ori) & 16.83$^\circ$ & 67.20\% & 0.8742 & 26.78$^\circ$ \\
Ours (gen. ori + opt. pos) & \textbf{7.53$^\circ$} & \textbf{83.87\%} & \textbf{0.9672} & \textbf{10.27$^\circ$} \\
\bottomrule
\end{tabular}
\end{center}
\end{table}

\section{Conclusions}
We presented CosmoH2G, a data-driven hand-to-gripper solution designed for object manipulations with complex spatial movements. At the data level, we introduced a scalable acquisition pipeline with a rigorous protocol that prioritizes motion complexity. This effort resulted in a large-scale hand-gripper paired dataset comprising 6,189 episodes across 1,254 unique objects, significantly surpassing existing benchmarks in manipulation complexity. Furthermore, we adopted a two-stage learning framework that simplifies the transfer process by predicting keyframe poses before generating continuous action sequences. Experiments demonstrate that our approach enables precise and stable hand-to-gripper transfer of complex spatial manipulations. 

\section{Limitations and Future Work}
While our two-stage framework effectively handles spatial transfer, unified one-stage direct transfer becomes increasingly viable as paired data scales. We aim to expand our dataset via the proposed scalable pipeline and refine the architecture for better generalizability.
Additionally, our framework currently operates open-loop, lacking real-time error correction and explicit collision avoidance. Integrating closed-loop feedback and collision-aware motion planning is a key direction for future work.
We believe our dataset and framework will provide a valuable foundation for future research in data-driven cross-embodiment learning and complex robotic manipulation.

\begin{acks}

The work was supported in part by Guangdong S\&T Programme with Grant No. 2024B0101030002，the Basic Research Project No. HZQB-KCZYZ-2021067 of Hetao Shenzhen-HK S\&T Cooperation Zone, the NSFC with Grant No. 62293482, Guangdong Provincial Fund for Distinguished Young Scholars No. 2023B1515020055, the Shenzhen Outstanding Talents Training Fund 202002, the Guangdong Provincial Key Laboratory of Future Networks of Intelligence (Grant No. 2022B1212010001), the Shenzhen Key Laboratory of Big Data and Artificial Intelligence (Grant No. SYSPG20241211173853027) , the Guangdong Province Radio Science Data Center with grant No. 2025B1212070001, the National Key R\&D Program of China with grant No. 2018YFB1800800.

\end{acks}

\clearpage

\bibliographystyle{ACM-Reference-Format}
\bibliography{sample-bibliography}


%
%
%
%

\clearpage
\appendix

\begin{figure*}
  \centering             
  \includegraphics[width=0.85\linewidth, keepaspectratio]{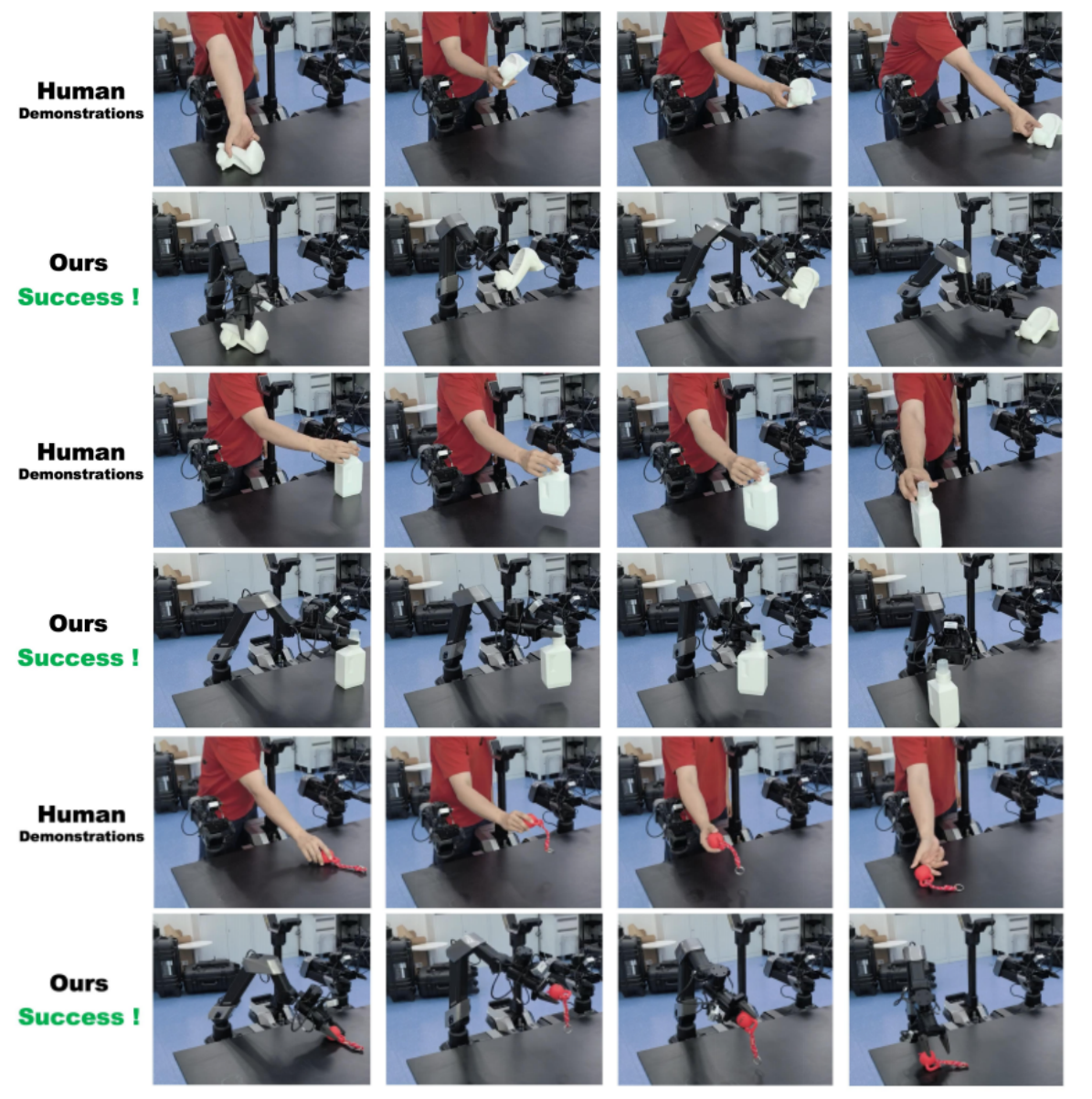}
  \caption{More results in real-robot experiments.}
  \label{fig:more-real-results}
\end{figure*}

\section{Implementation Details}

\subsection{Visualization of Registering and Tracking}
\label{sec:register-vis}

\begin{figure}[H]
  \centering             
  \includegraphics[width=0.9\linewidth, keepaspectratio]{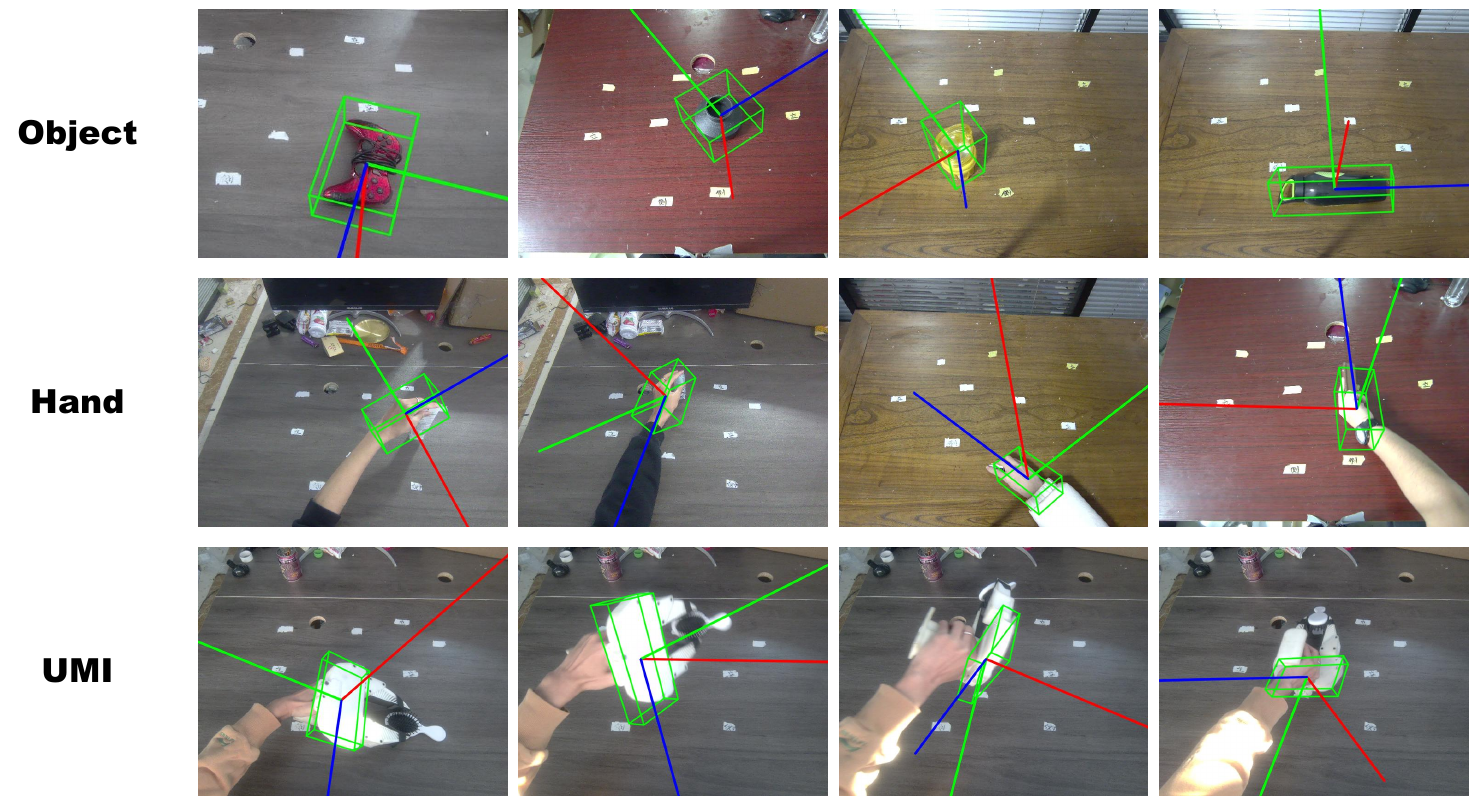}
  \caption{Visualization of registering and tracking. In UMI registration, we directly register the UMI base into the 3D scene, and its resulting 6-DOF pose is treated as the gripper pose.
}
\end{figure}

\subsection{Visualization of Gripper Keypoints}
\label{sec:keypoint-vis}

\begin{figure}[H]
  \centering             
  \includegraphics[width=0.5\linewidth, keepaspectratio]{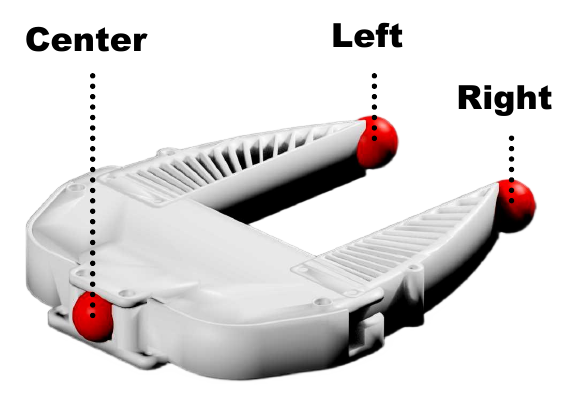}
  \caption{Visualization of gripper keypoints.
}
\end{figure}

\subsection{Hand-Object-Interaction Contact Map}
\label{sec:contactmap}

Given the hand point cloud $\{p^h_i\}_{i=1}^{N_h}$ and object point cloud $\{p^o_j\}_{j=1}^{N_o}$ in the shared reference frame, we compute a per-point \emph{HOI contact map} $m$ on the object surface. To this end, we first construct a KD-tree over the hand points and query the nearest-neighbor distance for each object point $p^o_j$, yielding $d_j = \min_{1\le i\le N_h} \| p^h_i - p^o_j \|_2$.
We then convert this distance into a normalized soft contact likelihood using a predefined maximum interaction radius $d_{\max}$ (e.g., derived from the typical finger contact range): $m_j \;=\; \mathrm{clip}\!\left( 1 - \frac{d_j}{d_{\max}},\, 0,\, 1 \right).$
where $m_j = 1$ indicates direct contact and the value linearly decays to $0$ beyond $d_{\max}$. Finally, we extract the set of \emph{valid contact points} $\mathcal{V}$ by thresholding the contact map at a fixed value $\tau$ to retain only high-confidence candidates: $\mathcal{V} = \bigl\{ p^o_j \mid m_j > \tau \bigr\}.$

\subsection{Optimization of Position Sequence}

\paragraph{\textbf{Initialization.}}
During inference, the gripper position sequence is computed from the given hand sequence. Under the prior that the contact regions are spatially aligned between hand and gripper, we use the contact region center as an anchor to calculate gripper positions. We first define the contact region centers. \textbf{Hand contact region center:} a hand point is classified as part of the contact region if its minimum distance to object points falls below a predefined \textit{contact threshold}:
\begin{equation}
\label{eqn:hand-contact-center}
d^h = \min_{\mathbf{p}^o \in \mathcal{O}} \|\mathbf{p}^h - \mathbf{p}^o\|,
\end{equation}
where $d^h$ denotes the minimum distance from hand point $\mathbf{p}^h$ to the object points, and $\mathcal{O}$  represents the set of object points. The hand contact center is then computed as the centroid of these contact points. \textbf{Gripper contact region center}: for the gripper, we define the midpoint between left and right finger keypoints as the contact region center. Although this may not be perfectly accurate, it provides a good initialization and will be refined through subsequent optimization. Then, we assume the hand and gripper contact region centers are aligned, and derive the gripper position sequence accordingly.

\paragraph{\textbf{Grasping Optimization.}}
We optimize the gripper positions at the starting and terminal frames through:

\begin{equation}
\label{eqn:contact-optimization-detail}
\Delta t^* = \arg \min_{\Delta t} \Big[ L_{\text{contact}}(\Delta t) + \lambda_{\Delta} L_{\text{reg}}(\Delta t) + \lambda_{\text{pen}} L_{\text{pen}}(\Delta t) \Big],
\end{equation}
where
\begin{align}
L_{\text{contact}}(\Delta t) &= \sum_{k \in \{L,R\}} \min_{\mathbf{v} \in \mathcal{V}} \big| (\mathbf{p}_k(\Delta t) - \mathbf{v}) \cdot \mathbf{d} \big|, \\
L_{\text{reg}}(\Delta t) &= \frac{1}{3} \| \Delta t \|_2^2, \\
L_{\text{pen}}(\Delta t) &= \sum_{\mathbf{g} \in \mathcal{G}(\Delta t)} \max\big(0, -d_{\text{signed}}(\mathbf{g}, \mathcal{O})\big)^2.
\end{align}
$\Delta t \in \mathbb{R}^3$ denotes the gripper position correction; $\mathbf{p}_L(\Delta t)$ and $\mathbf{p}_R(\Delta t)$ are the absolute coordinates of the left and right finger keypoints. $\mathcal{V}$ denotes the set of contacted object points (identified by the contact map), and $\mathbf{d}$ is the unit direction vector from $\mathbf{p}_L(\Delta t)$ to $\mathbf{p}_R(\Delta t)$. The first term, $L_{\text{contact}}$, penalizes the perpendicular distance from each finger keypoint to the contact points in $\mathcal{V}$, encouraging the gripper to approach and align with the hand contact region.
\textit{The second term} $L_{\text{reg}}$ penalizes large position corrections for a single optimization step via an L2 regularizer.
$\mathcal{G}(\Delta t)$ denotes the gripper point cloud after applying the position correction $\Delta t$, and $d_{\text{signed}}(\mathbf{x}, \mathcal{O})$ is the signed distance from point $\mathbf{x}$ to the object surface $\mathcal{O}$ (positive outside, negative inside). The third term, $L_{\text{pen}}$, penalizes any gripper point that falls inside the object (i.e., where $d_{\text{signed}}(\mathbf{g}, \mathcal{O}) < 0$), thereby preventing collision between the entire gripper body and the object, and ensuring physically feasible grasp configurations. Weights $\lambda_{\Delta}$ and $\lambda_{\text{pen}}$ balance the regularization on position correction against the penetration penalty.

\paragraph{Inverse Kinematics Optimization.} 
To ensure the optimized gripper poses are physically executable, we apply inverse kinematics (IK) optimization. Specifically, for each frame, we query an IK solver with the current 6-DoF pose and evaluate the end-effector residual and joint-limit violations. The 6-DoF pose is then adjusted to minimize this IK loss, ensuring the resulting configuration is both accurate and physically feasible. The objective is formulated as:

\begin{equation}
\label{eqn:ik-aware-detail}
a_i^* = \arg \min_{a_i} \Big[ L_{\text{ik}}(a_i) + \lambda_{\text{reg}} \| a_i - \hat{a}_i \|_2^2 \Big],
\end{equation}
where
\begin{align}
L_{\text{ik}}(a_i) &= \| \text{FK}_{\text{pos}}(q_i^*) - \mathbf{t}_i \|_2^2 + \lambda_{\text{ori}} \| \text{Log}(\mathbf{R}_i^\top \text{FK}_{\text{rot}}(q_i^*)) \|_2^2 + L_{\text{joint}}(q_i^*), \\
q_i^* &= \text{IK}(a_i), \\
L_{\text{joint}}(q_i^*) &= \sum_{j=1}^{J} \max\big(0, q_{i,j}^* - \bar{q}_j, \underline{q}_j - q_{i,j}^*\big)^2.
\end{align}
$a_i$ is the 6-DoF gripper pose to be optimized, and $\hat{a}_i$ is the initial pose from the previous stage. $q_i^* = \text{IK}(a_i)$ denotes the joint angles returned by the IK solver given pose target $a_i$. Thus, \textit{the first term} $L_{\text{ik}}$ penalizes the end-effector residual between the IK solution and the target pose, as well as joint-limit violations. $\text{FK}_{\text{pos}}(\cdot)$ and $\text{FK}_{\text{rot}}(\cdot)$ map joint angles to end-effector position and orientation, respectively. $\bar{q}_j$ and $\underline{q}_j$ denote the upper and lower bounds of the $j$-th joint. \textit{The second term} $\| a_i - \hat{a}_i \|_2^2$ regularizes the optimized pose to stay close to the original pose. $\lambda_{\text{reg}}$ balances IK feasibility and pose generation distribution.

\subsection{Real-robot Setup.}
\label{sec:setup-detail}

\begin{figure}[H]
  \centering             
  \includegraphics[width=0.7\linewidth, keepaspectratio]{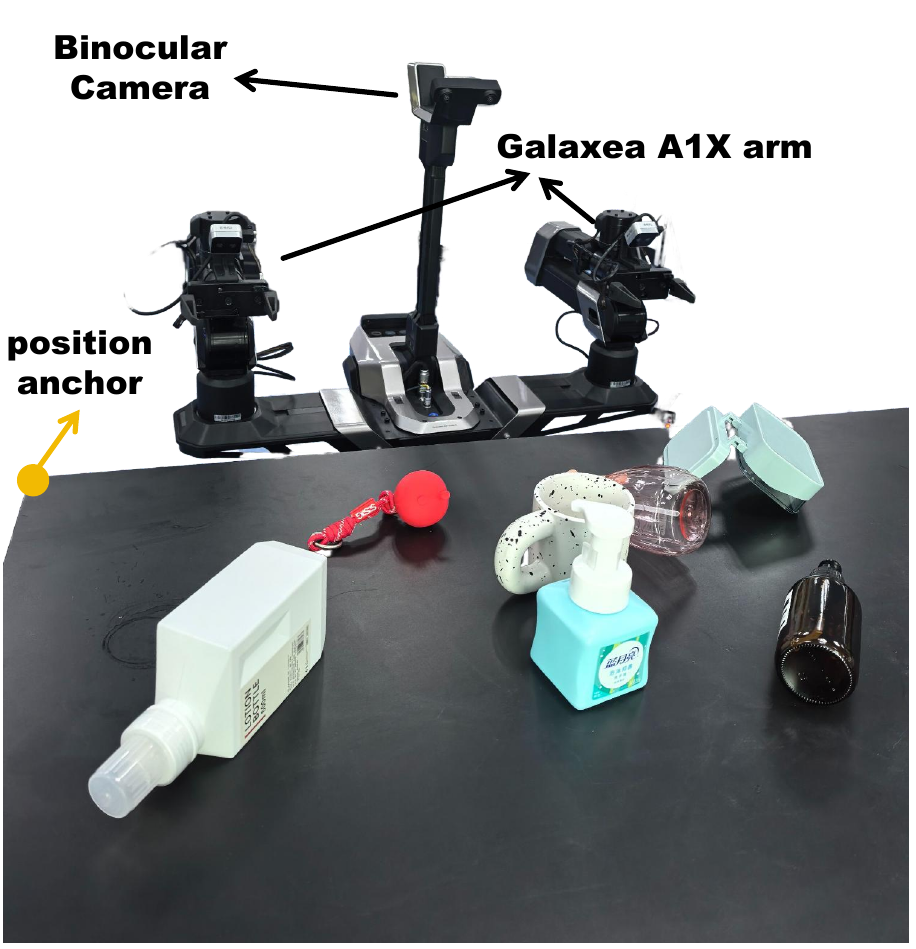}
  \caption{\textbf{Real-robot experimental setup.} We deploy our framework on a Galaxea R1 Lite equipped with two Galaxea A1X arms. A binocular camera is mounted at a fixed viewpoint to capture the workspace and human demonstrations. During evaluation, the object position is computed relative to the position anchor (yellow circle), and its orientation is estimated via FoundationPose++~\cite{foundationposeplusplus}.}
  \label{fig:setup}
\end{figure}

Our real-world evaluation setup uses a Galaxea R1 Lite equipped with two 7-DoF A1X arms and parallel-jaw G1 grippers. The world coordinate frame is defined on the tabletop, with its origin located at the table corner (indicated by the yellow circle in Fig.~\ref{fig:setup}), the $x$-axis pointing forward along the table edge, the $y$-axis pointing rightward, and the $z$-axis pointing upward.
The robot base is centered at this origin, and the arms are positioned symmetrically on either side. A binocular camera is rigidly mounted $0.4\,\text{m}$ above the origin, facing downward toward the table center to capture the manipulation workspace. The camera is extrinsically calibrated to the robot base frame prior to each data collection session. To ensure a fair comparison, the robot starts from a fixed initial configuration in all trials. During metric computation, the object position is calculated relative to the poisition anchor, while the object orientation is predicted by FoundationPose++~\cite{foundationposeplusplus}.

\subsection{Simulation Setup.}

Across evaluation, we maintain consistent object physics on the simulation platform. Each object's centroid is anchored at its mesh geometric center, with mass implicitly determined by a uniform density of approximately 1000 kg/m³ (representative of common plastic materials). Consistent with GraspVLA~\cite{deng2025graspvlagraspingfoundationmodel}, we further employ a low friction coefficient of 0.15 for all contact interactions. This standardized physical configuration aligns with the setup widely adopted in prior works including GraspVLA~\cite{deng2025graspvlagraspingfoundationmodel} and Im2Flow2Act~\cite{xu2024flowcrossdomainmanipulationinterface}.

\section{Baseline Details}
\label{baseline-detail}

For optimization-based methods, we select \textbf{MimicFunc}~\cite{tang2025mimicfuncimitatingtoolmanipulation} and \textbf{3DFlowAction}~\cite{zhi20253dflowactionlearningcrossembodimentmanipulation} as baselines. They first sample an initial 6-DOF grasping pose using GraspNet~\cite{fang2020graspnet} or AnyGrasp~\cite{fang2023anygrasprobustefficientgrasp},and then calculate the full sequence by applying object transformations. We exclude rule-based retargeting methods from our comparison because they are primarily designed for pinch gestures. As shown in Fig.~\ref{fig:phantom}, this limitation results in consistent failure in our task.

To provide a fair comparison, we retrain \textbf{Track2Act} on our dataset using 3D flows of hands as input. Specifically, we replace its original 2D tracks with a 3D hand point sequence and compute the initial action sequence via rule-based hand‑pose retargeting. Despite being provided with this enriched 3D information—arguably exceeding its original design assumptions—Track2Act still underperforms our CosmoH2G, underscoring the limitations of its underlying framework.

\section{Details for the Computed Variant}
\label{sec:computed-detail}

In \textit{Computed} variant, we construct a baseline without generation, that derives the gripper pose sequence through computation and optimization. 
In this baseline, directly using the demonstrated hand pose as the optimization target for gripper orientation is fundamentally unreliable because of their different morphologies, kinematic structures, and contact topologies.
We therefore reformulate by introducing the manipulated object as an intermediate geometric anchor. Its 6D pose sequence serves as a shared, task-agnostic reference: the hand drives the object's motion, and the gripper is tasked with replicating it.

\paragraph{Object Pose Induction via Hand Mesh Sequence.}
Direct visual tracking of the object is infeasible throughout the manipulation because the human hand persistently occludes the object. We therefore reconstruct the full hand mesh sequence and leverage the rigid contact relationship established at the initial frame to induce the object 6D pose trajectory. Concretely, at the first frame we obtain the initial object pose $O_1$, and extract the \textit{hand contact pose} $C_1$ from the reconstructed hand mesh. We then compute the constant relative transform $T_{c}^{o} = (O_1)^{-1} C_1$. Under the stable-grasp prior that the hand-object relative pose remains rigid during manipulation, the object pose at frame $t$ is induced as
\begin{equation}
O_t = C_t \, T_{c}^{o} = C_t \, C_1^{-1} O_1,
\end{equation}
where $C_t$ is the hand contact pose propagated from the reconstructed mesh at frame $t$. The resulting object trajectory $\{O_t\}_{t=1}^T$ subsequently serves as the unified optimization target for the gripper pose sequence.

\paragraph{Calculation of Position and Orientation Sequence.}
Before optimization, we derive the full gripper pose sequence in two stages.

\textbf{Boundary pose initialization.}
At the starting and terminal frames, we sample $K$ candidate 6-DoF grasp poses $\{G^{(k)}\}_{k=1}^K$ from the object point cloud via GraspNet~\cite{fang2020graspnet}. Each candidate is scored by three criteria: (\textit{i})~\textbf{contact-region consistency}---we compute the overlap between the candidate gripper's finger contact regions and the demonstrated contact map; (\textit{ii})~\textbf{orientation alignment}---the angular deviation between the candidate's approach axis and the hand's functional direction (i.e., the palm normal at the contact center); and (\textit{iii})~\textbf{grasp quality}---the confidence score from GraspNet. The top-scoring candidate is selected as the initial boundary pose.

\textbf{Intermediate-frame synthesis.}
The position sequence is computed identically to our proposed method. 
For orientation, we directly interpolate between the starting and terminal orientations via spherical linear interpolation (SLERP) on $SO(3)$.

\paragraph{Optimization of Full Pose Sequence.}
Given the initial position sequence, we refine it through contact-map alignment at the start and terminal frames (Eq.~\ref{eq:keyframe-opt}), temporal smoothness regularization (Eq.~\ref{eq:traj-opt}), identical to our proposed method.
For the orientation component, the boundary poses derived from GraspNet voting are held fixed, and the intermediate frames are refined under three objectives. First, \textbf{angular smoothness} is enforced by penalizing the second-order geodesic difference on $SO(3)$:
\begin{equation}
L_{\text{ori\_smooth}} = \sum_{i=1}^{T-2} \big\| \text{Log}(R_{i-1}^\top R_i) - \text{Log}(R_i^\top R_{i+1}) \big\|_2^2,
\end{equation}
where $\text{Log}(\cdot)$ maps the relative rotation to its axis-angle representation, discouraging abrupt motions. Second, the full gripper pose sequence is jointly optimized such that the object pose induced by the gripper aligns with the target object trajectory. The tracking loss penalizes the deviation between them:
\begin{equation}
L_{\text{track}} = \sum_{i=1}^{T-2} \left( \left\| \mathbf{t}_i^{o,\text{target}} - \tilde{\mathbf{t}}_i^o \right\|_2^2 + \lambda_{\text{ori}} \left\| \operatorname{Log}\left( (R_i^{o,\text{target}})^\top \tilde{R}_i^o \right) \right\|_2^2 \right),
\end{equation}
where $\tilde{\mathbf{t}}_i^o = \mathbf{t}_i^g - R_i^g \overline{\mathbf{d}}_{rel}$ and $\tilde{R}_i^o = R_i^g \overline{R}_{rel}^\top$ are the object position and orientation reconstructed from the current gripper pose $(R_i^g, \mathbf{t}_i^g)$ using the average relative transform $(\overline{R}_{rel}, \overline{\mathbf{d}}_{rel})$ observed at the boundary frames. This loss encourages the gripper to achieve poses whose derived object pose (under the fixed relative transform) is consistent with the demonstrated trajectory. Third, the \textbf{IK stage} jointly projects both position and orientation into the robot's feasible joint space via the full residual in Eq.~\ref{eq:ik-opt}.

\paragraph{Failure Cases.} However, the \textit{Computed} variant always fails on tasks involving complex spatial motions such as flipping. First, the SLERP-interpolated initialization provides only a rough starting guess. The intermediate orientation is far from the true rotation, leaving the optimizer lost in a large search space. Second, jointly optimizing position and orientation creates many equally valid but drastically different solutions. This ambiguity often breaks temporal continuity, causing abrupt jumps or unnatural motions. Together, these issues show that accurate orientation trajectories must instead be learned directly from data.

\section{Evaluation Metrics}
\label{sec:metric}
This section presents the formal definitions and implementation details of our evaluation metrics.

\subsection{Trajectory Similarity}
\label{sec:TS}
Trajectory similarity is quantitatively assessed by comparing the 3D position trajectories between ground truth (GT) and the executed paths. 
Specifically, we compute the \textit{Dynamic Time Warping (DTW)} distance between them using the FastDTW algorithm~\cite{wu2020fastdtw}, and normalize it by sequence length.
To convert raw distances into an interpretable similarity score $S \in [0, 1]$, we normalize the metric distance $D_{\text{metric}}$ by a task-specific tolerance threshold $\tau$ (set as 20\% of the table width):
\[
S = 1 - \min\left(\frac{D_{\text{metric}}}{\tau}, 1\right).
\]
Under this formulation, deviations within the tolerance bound yield proportionally decreasing similarity, while any deviation exceeding $\tau$ is clamped to zero similarity.

\paragraph{Implementation details.} To compute the hand positional trajectory, we use the contact region center of each reconstructed hand mesh as the position point and concatenate these points across the sequence. Similarly, for the gripper positional trajectory, we adopt the contact region center of end-effector as the position point. On the real robot, this trajectory is obtained from sensor feedback (joint encoders and forward kinematics) during execution. In simulation, the trajectory is directly queried from the physics engine via the end-effector link state at each timestep.

\subsection{Success Rate.}
\label{sec:SR}
Following prior work~\cite{xu2024flowcrossdomainmanipulationinterface}, we adopt the success rate to assess position-level task success, defined by the distance between the final placement and the desired target.
Let \( W \) denote the width of the desktop, and let \( d \) represent the Euclidean distance between the final placed position and the target. A placement is considered a \textbf{failure} if the distance exceeds 10\% of the desktop width. For a series of \( N \) placement trials, the success rate \( R \) is calculated as:

\[
R = \frac{N_{\text{success}}}{N} \times 100\%,
\]

where \( N_{\text{success}} \) is the number of trials satisfying:

\[
d \leq 0.1 \times W.
\]

\subsection{Target orientation placement accuracy.}
We adopt the Target Orientation Placement Accuracy (TOPA) to assess orientation-level placement precision, defined by the angular deviation between the final placed orientation and the desired target orientation. Let $\theta$ denote the angular error between the final orientation and the target orientation, measured in degrees. For a series of $N$ placement trials, the TOPA score is calculated as the mean angular error:

\[
\text{TOPA} = \frac{1}{N} \sum_{i=1}^{N} \theta_i \quad (\text{in degrees}),
\]

where $\theta_i$ represents the angular error of the $i$-th trial. In real-robot experiments, object orientations are estimated via FoundationPose++~\cite{foundationposeplusplus}.

\begin{figure}[H]
  \centering             
  \includegraphics[width=0.7\linewidth, keepaspectratio]{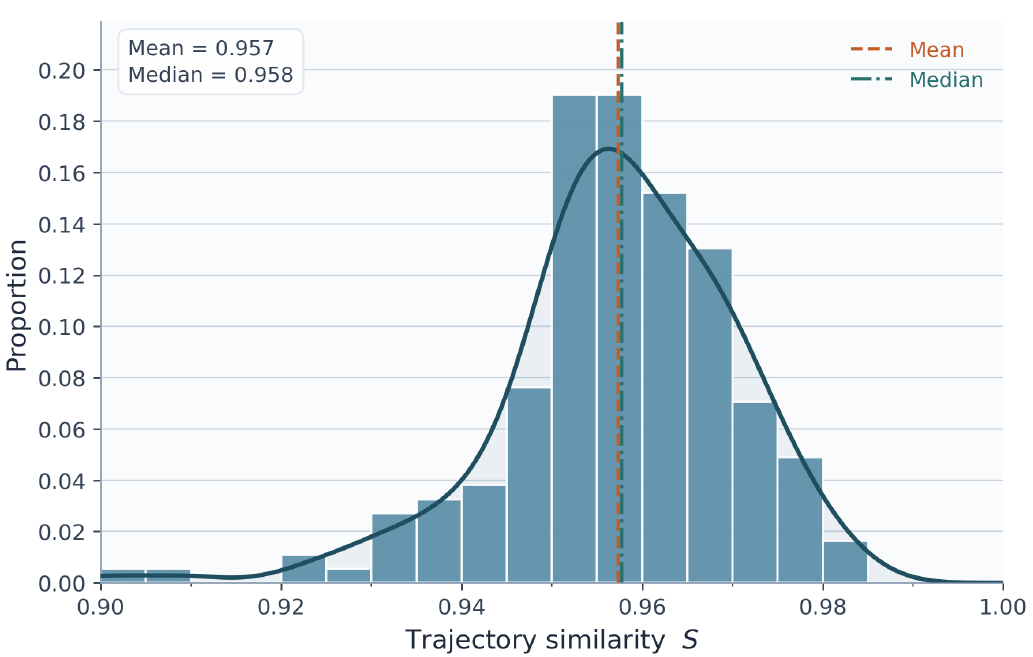}
  \caption{\textbf{Similarity distribution of retained hand-UMI pairs.} The similarity scores (mean = 0.957, median = 0.958) concentrate well above the 0.9 retention threshold.}
  \label{fig:TS}
\end{figure}

\section{Generalization Analysis}
\label{sec:generalization}

Beyond novel objects and motions, we further assess the 
generalization scope of CosmoH2G along  demonstrators and camera viewpoints.

\paragraph{\textbf{Demonstrators.}} Our dataset is collected from multiple male and female participants. To further examine hand-scale generalization, we construct an additional test split consisting of 4 adult males, 4 adult females, and 2 children. As shown in Table~\ref{tab:hand-scale}, CosmoH2G generalizes well across adult hands of varying scales (SR of 83.33\% and 91.67\% for adult males and females, respectively, comparable to 83.87\% on the original test set). Performance degrades on children's hands (SR 33.33\%), whose scales fall outside the training distribution, indicating the boundary of hand-scale generalization.

\begin{table}[t]
    \centering
    \caption{\textbf{Generalization across Demonstrators (in simulation).} The model generalizes well across adult hands, while performance degrades on children's hands, whose scales fall outside the training distribution.}
    \label{tab:hand-scale}
    \resizebox{\linewidth}{!}{
    \begin{tabular}{lcccc}
        \toprule
        Test set & Original & Adult males & Adult females & Children \\
        \midrule
        GOA $\downarrow$ & 7.53$^\circ$ & \textbf{6.74$^\circ$} & 7.89$^\circ$ & 18.67$^\circ$ \\
        SR $\uparrow$ & 83.87\% & 83.33\% & \textbf{91.67\%} & 33.33\% \\
        TS $\uparrow$ & 0.9672 & 0.9638 & \textbf{0.9728} & 0.8215 \\
        TOPA $\downarrow$ & 10.27$^\circ$ & 12.32$^\circ$ & \textbf{9.93$^\circ$} & 51.80$^\circ$ \\
        \bottomrule
    \end{tabular}
    }
\end{table}

\paragraph{\textbf{Camera Viewpoints.}} Since our input is 3D data 
extracted from videos, CosmoH2G is insensitive to the four evaluated camera viewpoints. We collect 20 unseen cases recorded from four distinct viewpoints, including first-person, egocentric top-down, third-person, and exocentric top-down. Performance remains stable across all viewpoints (SR ranging from 75\% to 90\%), as shown in Table~\ref{tab:viewpoint}.

\begin{table}[t]
    \centering
    \caption{\textbf{Generalization across camera viewpoints (in simulation).} Since our input is 3D data extracted from videos, performance remains stable across all viewpoints.}
    \label{tab:viewpoint}
    \resizebox{\linewidth}{!}{
    \begin{tabular}{lcccc}
        \toprule
        Viewpoint & First-person & Ego. top-down & Third-person & Exo. top-down \\
        \midrule
        GOA (sim) $\downarrow$ & 6.73$^\circ$ & \textbf{6.21$^\circ$} & 9.58$^\circ$ & 7.94$^\circ$ \\
        SR (sim) $\uparrow$ & \textbf{90\%} & 85\% & 75\% & \textbf{90\%} \\
        TS (sim) $\uparrow$ & 0.9617 & 0.9479 & 0.9032 & \textbf{0.9681} \\
        TOPA (sim) $\downarrow$ & 11.02$^\circ$ & 10.73$^\circ$ & 19.17$^\circ$ & \textbf{10.49$^\circ$} \\
        \bottomrule
    \end{tabular}
    }
\end{table}

%% file: sample-acmtog-SIGGRAPH-submission.bbl

\begin{thebibliography}{55}


\ifx \showCODEN    \undefined \def \showCODEN     #1{\unskip}     \fi
\ifx \showISBNx    \undefined \def \showISBNx     #1{\unskip}     \fi
\ifx \showISBNxiii \undefined \def \showISBNxiii  #1{\unskip}     \fi
\ifx \showISSN     \undefined \def \showISSN      #1{\unskip}     \fi
\ifx \showLCCN     \undefined \def \showLCCN      #1{\unskip}     \fi
\ifx \shownote     \undefined \def \shownote      #1{#1}          \fi
\ifx \showarticletitle \undefined \def \showarticletitle #1{#1}   \fi
\ifx \showURL      \undefined \def \showURL       {\relax}        \fi
\providecommand\bibfield[2]{#2}
\providecommand\bibinfo[2]{#2}
\providecommand\natexlab[1]{#1}
\providecommand\showeprint[2][]{arXiv:#2}

\bibitem[Bharadhwaj et~al\mbox{.}(2024a)]%
        {bharadhwaj2024gen2acthumanvideogeneration}
\bibfield{author}{\bibinfo{person}{Homanga Bharadhwaj}, \bibinfo{person}{Debidatta Dwibedi}, \bibinfo{person}{Abhinav Gupta}, \bibinfo{person}{Shubham Tulsiani}, \bibinfo{person}{Carl Doersch}, \bibinfo{person}{Ted Xiao}, \bibinfo{person}{Dhruv Shah}, \bibinfo{person}{Fei Xia}, \bibinfo{person}{Dorsa Sadigh}, {and} \bibinfo{person}{Sean Kirmani}.} \bibinfo{year}{2024}\natexlab{a}.
\newblock \bibinfo{title}{Gen2Act: Human Video Generation in Novel Scenarios enables Generalizable Robot Manipulation}.
\newblock
\showeprint[arxiv]{2409.16283}~[cs.RO]
\urldef\tempurl%
\url{https://arxiv.org/abs/2409.16283}
\showURL{%
\tempurl}


\bibitem[Bharadhwaj et~al\mbox{.}(2024b)]%
        {bharadhwaj2024track2actpredictingpointtracks}
\bibfield{author}{\bibinfo{person}{Homanga Bharadhwaj}, \bibinfo{person}{Roozbeh Mottaghi}, \bibinfo{person}{Abhinav Gupta}, {and} \bibinfo{person}{Shubham Tulsiani}.} \bibinfo{year}{2024}\natexlab{b}.
\newblock \bibinfo{title}{Track2Act: Predicting Point Tracks from Internet Videos enables Generalizable Robot Manipulation}.
\newblock
\showeprint[arxiv]{2405.01527}~[cs.RO]
\urldef\tempurl%
\url{https://arxiv.org/abs/2405.01527}
\showURL{%
\tempurl}


\bibitem[Brohan et~al\mbox{.}(2023)]%
        {brohan2023rt1roboticstransformerrealworld}
\bibfield{author}{\bibinfo{person}{Anthony Brohan}, \bibinfo{person}{Noah Brown}, \bibinfo{person}{Justice Carbajal}, \bibinfo{person}{Yevgen Chebotar}, \bibinfo{person}{Joseph Dabis}, \bibinfo{person}{Chelsea Finn}, \bibinfo{person}{Keerthana Gopalakrishnan}, \bibinfo{person}{Karol Hausman}, \bibinfo{person}{Alex Herzog}, \bibinfo{person}{Jasmine Hsu}, \bibinfo{person}{Julian Ibarz}, \bibinfo{person}{Brian Ichter}, \bibinfo{person}{Alex Irpan}, \bibinfo{person}{Tomas Jackson}, \bibinfo{person}{Sally Jesmonth}, \bibinfo{person}{Nikhil~J Joshi}, \bibinfo{person}{Ryan Julian}, \bibinfo{person}{Dmitry Kalashnikov}, \bibinfo{person}{Yuheng Kuang}, \bibinfo{person}{Isabel Leal}, \bibinfo{person}{Kuang-Huei Lee}, \bibinfo{person}{Sergey Levine}, \bibinfo{person}{Yao Lu}, \bibinfo{person}{Utsav Malla}, \bibinfo{person}{Deeksha Manjunath}, \bibinfo{person}{Igor Mordatch}, \bibinfo{person}{Ofir Nachum}, \bibinfo{person}{Carolina Parada}, \bibinfo{person}{Jodilyn Peralta}, \bibinfo{person}{Emily Perez},
  \bibinfo{person}{Karl Pertsch}, \bibinfo{person}{Jornell Quiambao}, \bibinfo{person}{Kanishka Rao}, \bibinfo{person}{Michael Ryoo}, \bibinfo{person}{Grecia Salazar}, \bibinfo{person}{Pannag Sanketi}, \bibinfo{person}{Kevin Sayed}, \bibinfo{person}{Jaspiar Singh}, \bibinfo{person}{Sumedh Sontakke}, \bibinfo{person}{Austin Stone}, \bibinfo{person}{Clayton Tan}, \bibinfo{person}{Huong Tran}, \bibinfo{person}{Vincent Vanhoucke}, \bibinfo{person}{Steve Vega}, \bibinfo{person}{Quan Vuong}, \bibinfo{person}{Fei Xia}, \bibinfo{person}{Ted Xiao}, \bibinfo{person}{Peng Xu}, \bibinfo{person}{Sichun Xu}, \bibinfo{person}{Tianhe Yu}, {and} \bibinfo{person}{Brianna Zitkovich}.} \bibinfo{year}{2023}\natexlab{}.
\newblock \bibinfo{title}{RT-1: Robotics Transformer for Real-World Control at Scale}.
\newblock
\showeprint[arxiv]{2212.06817}~[cs.RO]
\urldef\tempurl%
\url{https://arxiv.org/abs/2212.06817}
\showURL{%
\tempurl}


\bibitem[Cai et~al\mbox{.}(2024)]%
        {cai2024visualimitationlearningtaskoriented}
\bibfield{author}{\bibinfo{person}{Yichen Cai}, \bibinfo{person}{Jianfeng Gao}, \bibinfo{person}{Christoph Pohl}, {and} \bibinfo{person}{Tamim Asfour}.} \bibinfo{year}{2024}\natexlab{}.
\newblock \bibinfo{title}{Visual Imitation Learning of Task-Oriented Object Grasping and Rearrangement}.
\newblock
\showeprint[arxiv]{2403.14000}~[cs.RO]
\urldef\tempurl%
\url{https://arxiv.org/abs/2403.14000}
\showURL{%
\tempurl}


\bibitem[Chang et~al\mbox{.}(2025)]%
        {chang2025reconviagenaccuratemultiview3d}
\bibfield{author}{\bibinfo{person}{Jiahao Chang}, \bibinfo{person}{Chongjie Ye}, \bibinfo{person}{Yushuang Wu}, \bibinfo{person}{Yuantao Chen}, \bibinfo{person}{Yidan Zhang}, \bibinfo{person}{Zhongjin Luo}, \bibinfo{person}{Chenghong Li}, \bibinfo{person}{Yihao Zhi}, {and} \bibinfo{person}{Xiaoguang Han}.} \bibinfo{year}{2025}\natexlab{}.
\newblock \bibinfo{title}{ReconViaGen: Towards Accurate Multi-view 3D Object Reconstruction via Generation}.
\newblock
\showeprint[arxiv]{2510.23306}~[cs.CV]
\urldef\tempurl%
\url{https://arxiv.org/abs/2510.23306}
\showURL{%
\tempurl}


\bibitem[Chen et~al\mbox{.}(2025b)]%
        {chen2025toolasinterfacelearningrobotpolicies}
\bibfield{author}{\bibinfo{person}{Haonan Chen}, \bibinfo{person}{Cheng Zhu}, \bibinfo{person}{Shuijing Liu}, \bibinfo{person}{Yunzhu Li}, {and} \bibinfo{person}{Katherine Driggs-Campbell}.} \bibinfo{year}{2025}\natexlab{b}.
\newblock \bibinfo{title}{Tool-as-Interface: Learning Robot Policies from Observing Human Tool Use}.
\newblock
\showeprint[arxiv]{2504.04612}~[cs.RO]
\urldef\tempurl%
\url{https://arxiv.org/abs/2504.04612}
\showURL{%
\tempurl}


\bibitem[Chen et~al\mbox{.}(2025a)]%
        {chen2025videodepthanythingconsistent}
\bibfield{author}{\bibinfo{person}{Sili Chen}, \bibinfo{person}{Hengkai Guo}, \bibinfo{person}{Shengnan Zhu}, \bibinfo{person}{Feihu Zhang}, \bibinfo{person}{Zilong Huang}, \bibinfo{person}{Jiashi Feng}, {and} \bibinfo{person}{Bingyi Kang}.} \bibinfo{year}{2025}\natexlab{a}.
\newblock \bibinfo{title}{Video Depth Anything: Consistent Depth Estimation for Super-Long Videos}.
\newblock
\showeprint[arxiv]{2501.12375}~[cs.CV]
\urldef\tempurl%
\url{https://arxiv.org/abs/2501.12375}
\showURL{%
\tempurl}


\bibitem[Chi et~al\mbox{.}(2023)]%
        {chi2023diffusionpolicy}
\bibfield{author}{\bibinfo{person}{Cheng Chi}, \bibinfo{person}{Siyuan Feng}, \bibinfo{person}{Yilun Du}, \bibinfo{person}{Zhenjia Xu}, \bibinfo{person}{Eric Cousineau}, \bibinfo{person}{Benjamin Burchfiel}, {and} \bibinfo{person}{Shuran Song}.} \bibinfo{year}{2023}\natexlab{}.
\newblock \showarticletitle{Diffusion Policy: Visuomotor Policy Learning via Action Diffusion}. In \bibinfo{booktitle}{\emph{Proceedings of Robotics: Science and Systems (RSS)}}.
\newblock


\bibitem[Chi et~al\mbox{.}(2024)]%
        {chi2024universalmanipulationinterfaceinthewild}
\bibfield{author}{\bibinfo{person}{Cheng Chi}, \bibinfo{person}{Zhenjia Xu}, \bibinfo{person}{Chuer Pan}, \bibinfo{person}{Eric Cousineau}, \bibinfo{person}{Benjamin Burchfiel}, \bibinfo{person}{Siyuan Feng}, \bibinfo{person}{Russ Tedrake}, {and} \bibinfo{person}{Shuran Song}.} \bibinfo{year}{2024}\natexlab{}.
\newblock \bibinfo{title}{Universal Manipulation Interface: In-The-Wild Robot Teaching Without In-The-Wild Robots}.
\newblock
\showeprint[arxiv]{2402.10329}~[cs.RO]
\urldef\tempurl%
\url{https://arxiv.org/abs/2402.10329}
\showURL{%
\tempurl}


\bibitem[Deng et~al\mbox{.}(2025)]%
        {deng2025graspvlagraspingfoundationmodel}
\bibfield{author}{\bibinfo{person}{Shengliang Deng}, \bibinfo{person}{Mi Yan}, \bibinfo{person}{Songlin Wei}, \bibinfo{person}{Haixin Ma}, \bibinfo{person}{Yuxin Yang}, \bibinfo{person}{Jiayi Chen}, \bibinfo{person}{Zhiqi Zhang}, \bibinfo{person}{Taoyu Yang}, \bibinfo{person}{Xuheng Zhang}, \bibinfo{person}{Wenhao Zhang}, \bibinfo{person}{Heming Cui}, \bibinfo{person}{Zhizheng Zhang}, {and} \bibinfo{person}{He Wang}.} \bibinfo{year}{2025}\natexlab{}.
\newblock \bibinfo{title}{GraspVLA: a Grasping Foundation Model Pre-trained on Billion-scale Synthetic Action Data}.
\newblock
\showeprint[arxiv]{2505.03233}~[cs.RO]
\urldef\tempurl%
\url{https://arxiv.org/abs/2505.03233}
\showURL{%
\tempurl}


\bibitem[Dessalene et~al\mbox{.}(2025)]%
        {dessalene2025embodiswapzeroshotrobotimitation}
\bibfield{author}{\bibinfo{person}{Eadom Dessalene}, \bibinfo{person}{Pavan Mantripragada}, \bibinfo{person}{Michael Maynord}, {and} \bibinfo{person}{Yiannis Aloimonos}.} \bibinfo{year}{2025}\natexlab{}.
\newblock \bibinfo{title}{EmbodiSwap for Zero-Shot Robot Imitation Learning}.
\newblock
\showeprint[arxiv]{2510.03706}~[cs.RO]
\urldef\tempurl%
\url{https://arxiv.org/abs/2510.03706}
\showURL{%
\tempurl}


\bibitem[Ding et~al\mbox{.}(2024)]%
        {ding2024bunnyvisionprorealtimebimanualdexterous}
\bibfield{author}{\bibinfo{person}{Runyu Ding}, \bibinfo{person}{Yuzhe Qin}, \bibinfo{person}{Jiyue Zhu}, \bibinfo{person}{Chengzhe Jia}, \bibinfo{person}{Shiqi Yang}, \bibinfo{person}{Ruihan Yang}, \bibinfo{person}{Xiaojuan Qi}, {and} \bibinfo{person}{Xiaolong Wang}.} \bibinfo{year}{2024}\natexlab{}.
\newblock \bibinfo{title}{Bunny-VisionPro: Real-Time Bimanual Dexterous Teleoperation for Imitation Learning}.
\newblock
\showeprint[arxiv]{2407.03162}~[cs.RO]
\urldef\tempurl%
\url{https://arxiv.org/abs/2407.03162}
\showURL{%
\tempurl}


\bibitem[Dong et~al\mbox{.}(2024)]%
        {dong2024rtagrasplearningtaskorientedgrasping}
\bibfield{author}{\bibinfo{person}{Wenlong Dong}, \bibinfo{person}{Dehao Huang}, \bibinfo{person}{Jiangshan Liu}, \bibinfo{person}{Chao Tang}, {and} \bibinfo{person}{Hong Zhang}.} \bibinfo{year}{2024}\natexlab{}.
\newblock \bibinfo{title}{RTAGrasp: Learning Task-Oriented Grasping from Human Videos via Retrieval, Transfer, and Alignment}.
\newblock
\showeprint[arxiv]{2409.16033}~[cs.RO]
\urldef\tempurl%
\url{https://arxiv.org/abs/2409.16033}
\showURL{%
\tempurl}


\bibitem[Fang et~al\mbox{.}(2023)]%
        {fang2023anygrasprobustefficientgrasp}
\bibfield{author}{\bibinfo{person}{Hao-Shu Fang}, \bibinfo{person}{Chenxi Wang}, \bibinfo{person}{Hongjie Fang}, \bibinfo{person}{Minghao Gou}, \bibinfo{person}{Jirong Liu}, \bibinfo{person}{Hengxu Yan}, \bibinfo{person}{Wenhai Liu}, \bibinfo{person}{Yichen Xie}, {and} \bibinfo{person}{Cewu Lu}.} \bibinfo{year}{2023}\natexlab{}.
\newblock \bibinfo{title}{AnyGrasp: Robust and Efficient Grasp Perception in Spatial and Temporal Domains}.
\newblock
\showeprint[arxiv]{2212.08333}~[cs.RO]
\urldef\tempurl%
\url{https://arxiv.org/abs/2212.08333}
\showURL{%
\tempurl}


\bibitem[Fang et~al\mbox{.}(2020)]%
        {fang2020graspnet}
\bibfield{author}{\bibinfo{person}{Hao-Shu Fang}, \bibinfo{person}{Chenxi Wang}, \bibinfo{person}{Minghao Gou}, {and} \bibinfo{person}{Cewu Lu}.} \bibinfo{year}{2020}\natexlab{}.
\newblock \showarticletitle{GraspNet-1Billion: A Large-Scale Benchmark for General Object Grasping}. In \bibinfo{booktitle}{\emph{Proceedings of the IEEE/CVF Conference on Computer Vision and Pattern Recognition(CVPR)}}. \bibinfo{pages}{11444--11453}.
\newblock


\bibitem[Fu et~al\mbox{.}(2024)]%
        {fu2024mobilealohalearningbimanual}
\bibfield{author}{\bibinfo{person}{Zipeng Fu}, \bibinfo{person}{Tony~Z. Zhao}, {and} \bibinfo{person}{Chelsea Finn}.} \bibinfo{year}{2024}\natexlab{}.
\newblock \bibinfo{title}{Mobile ALOHA: Learning Bimanual Mobile Manipulation with Low-Cost Whole-Body Teleoperation}.
\newblock
\showeprint[arxiv]{2401.02117}~[cs.RO]
\urldef\tempurl%
\url{https://arxiv.org/abs/2401.02117}
\showURL{%
\tempurl}


\bibitem[Haldar and Pinto(2025)]%
        {haldar2025pointpolicyunifyingobservations}
\bibfield{author}{\bibinfo{person}{Siddhant Haldar} {and} \bibinfo{person}{Lerrel Pinto}.} \bibinfo{year}{2025}\natexlab{}.
\newblock \bibinfo{title}{Point Policy: Unifying Observations and Actions with Key Points for Robot Manipulation}.
\newblock
\showeprint[arxiv]{2502.20391}~[cs.RO]
\urldef\tempurl%
\url{https://arxiv.org/abs/2502.20391}
\showURL{%
\tempurl}


\bibitem[Heppert et~al\mbox{.}(2024)]%
        {Heppert_2024}
\bibfield{author}{\bibinfo{person}{Nick Heppert}, \bibinfo{person}{Max Argus}, \bibinfo{person}{Tim Welschehold}, \bibinfo{person}{Thomas Brox}, {and} \bibinfo{person}{Abhinav Valada}.} \bibinfo{year}{2024}\natexlab{}.
\newblock \showarticletitle{DITTO: Demonstration Imitation by Trajectory Transformation}. In \bibinfo{booktitle}{\emph{2024 IEEE/RSJ International Conference on Intelligent Robots and Systems (IROS)}}. \bibinfo{publisher}{IEEE}, \bibinfo{pages}{7565–7572}.
\newblock
\href{https://doi.org/10.1109/iros58592.2024.10801982}{doi:\nolinkurl{10.1109/iros58592.2024.10801982}}


\bibitem[Huang et~al\mbox{.}(2025a)]%
        {huang2025hgdiffuserefficienttaskorientedgrasp}
\bibfield{author}{\bibinfo{person}{Dehao Huang}, \bibinfo{person}{Wenlong Dong}, \bibinfo{person}{Chao Tang}, {and} \bibinfo{person}{Hong Zhang}.} \bibinfo{year}{2025}\natexlab{a}.
\newblock \bibinfo{title}{HGDiffuser: Efficient Task-Oriented Grasp Generation via Human-Guided Grasp Diffusion Models}.
\newblock
\showeprint[arxiv]{2503.00508}~[cs.RO]
\urldef\tempurl%
\url{https://arxiv.org/abs/2503.00508}
\showURL{%
\tempurl}


\bibitem[Huang et~al\mbox{.}(2025b)]%
        {huang2025umigenunifiedframeworkegocentric}
\bibfield{author}{\bibinfo{person}{Yan Huang}, \bibinfo{person}{Shoujie Li}, \bibinfo{person}{Xingting Li}, {and} \bibinfo{person}{Wenbo Ding}.} \bibinfo{year}{2025}\natexlab{b}.
\newblock \bibinfo{title}{UMIGen: A Unified Framework for Egocentric Point Cloud Generation and Cross-Embodiment Robotic Imitation Learning}.
\newblock
\showeprint[arxiv]{2511.09302}~[cs.RO]
\urldef\tempurl%
\url{https://arxiv.org/abs/2511.09302}
\showURL{%
\tempurl}


\bibitem[Iyer et~al\mbox{.}(2024)]%
        {iyer2024openteachversatileteleoperation}
\bibfield{author}{\bibinfo{person}{Aadhithya Iyer}, \bibinfo{person}{Zhuoran Peng}, \bibinfo{person}{Yinlong Dai}, \bibinfo{person}{Irmak Guzey}, \bibinfo{person}{Siddhant Haldar}, \bibinfo{person}{Soumith Chintala}, {and} \bibinfo{person}{Lerrel Pinto}.} \bibinfo{year}{2024}\natexlab{}.
\newblock \bibinfo{title}{OPEN TEACH: A Versatile Teleoperation System for Robotic Manipulation}.
\newblock
\showeprint[arxiv]{2403.07870}~[cs.RO]
\urldef\tempurl%
\url{https://arxiv.org/abs/2403.07870}
\showURL{%
\tempurl}


\bibitem[Ju et~al\mbox{.}(2024)]%
        {ju2024roboabcaffordancegeneralizationcategories}
\bibfield{author}{\bibinfo{person}{Yuanchen Ju}, \bibinfo{person}{Kaizhe Hu}, \bibinfo{person}{Guowei Zhang}, \bibinfo{person}{Gu Zhang}, \bibinfo{person}{Mingrun Jiang}, {and} \bibinfo{person}{Huazhe Xu}.} \bibinfo{year}{2024}\natexlab{}.
\newblock \bibinfo{title}{Robo-ABC: Affordance Generalization Beyond Categories via Semantic Correspondence for Robot Manipulation}.
\newblock
\showeprint[arxiv]{2401.07487}~[cs.RO]
\urldef\tempurl%
\url{https://arxiv.org/abs/2401.07487}
\showURL{%
\tempurl}


\bibitem[Lepert et~al\mbox{.}(2025a)]%
        {lepert2025masqueradelearninginthewildhuman}
\bibfield{author}{\bibinfo{person}{Marion Lepert}, \bibinfo{person}{Jiaying Fang}, {and} \bibinfo{person}{Jeannette Bohg}.} \bibinfo{year}{2025}\natexlab{a}.
\newblock \bibinfo{title}{Masquerade: Learning from In-the-wild Human Videos using Data-Editing}.
\newblock
\showeprint[arxiv]{2508.09976}~[cs.RO]
\urldef\tempurl%
\url{https://arxiv.org/abs/2508.09976}
\showURL{%
\tempurl}


\bibitem[Lepert et~al\mbox{.}(2025b)]%
        {lepert2025phantomtrainingrobotsrobots}
\bibfield{author}{\bibinfo{person}{Marion Lepert}, \bibinfo{person}{Jiaying Fang}, {and} \bibinfo{person}{Jeannette Bohg}.} \bibinfo{year}{2025}\natexlab{b}.
\newblock \bibinfo{title}{Phantom: Training Robots Without Robots Using Only Human Videos}.
\newblock
\showeprint[arxiv]{2503.00779}~[cs.RO]
\urldef\tempurl%
\url{https://arxiv.org/abs/2503.00779}
\showURL{%
\tempurl}


\bibitem[Mandlekar et~al\mbox{.}(2018)]%
        {mandlekar2018roboturkcrowdsourcingplatformrobotic}
\bibfield{author}{\bibinfo{person}{Ajay Mandlekar}, \bibinfo{person}{Yuke Zhu}, \bibinfo{person}{Animesh Garg}, \bibinfo{person}{Jonathan Booher}, \bibinfo{person}{Max Spero}, \bibinfo{person}{Albert Tung}, \bibinfo{person}{Julian Gao}, \bibinfo{person}{John Emmons}, \bibinfo{person}{Anchit Gupta}, \bibinfo{person}{Emre Orbay}, \bibinfo{person}{Silvio Savarese}, {and} \bibinfo{person}{Li Fei-Fei}.} \bibinfo{year}{2018}\natexlab{}.
\newblock \bibinfo{title}{RoboTurk: A Crowdsourcing Platform for Robotic Skill Learning through Imitation}.
\newblock
\showeprint[arxiv]{1811.02790}~[cs.RO]
\urldef\tempurl%
\url{https://arxiv.org/abs/1811.02790}
\showURL{%
\tempurl}


\bibitem[Park et~al\mbox{.}(2025)]%
        {park2025demodiffusiononeshothumanimitation}
\bibfield{author}{\bibinfo{person}{Sungjae Park}, \bibinfo{person}{Homanga Bharadhwaj}, {and} \bibinfo{person}{Shubham Tulsiani}.} \bibinfo{year}{2025}\natexlab{}.
\newblock \bibinfo{title}{DemoDiffusion: One-Shot Human Imitation using pre-trained Diffusion Policy}.
\newblock
\showeprint[arxiv]{2506.20668}~[cs.RO]
\urldef\tempurl%
\url{https://arxiv.org/abs/2506.20668}
\showURL{%
\tempurl}


\bibitem[Peebles and Xie(2023)]%
        {peebles2023scalablediffusionmodelstransformers}
\bibfield{author}{\bibinfo{person}{William Peebles} {and} \bibinfo{person}{Saining Xie}.} \bibinfo{year}{2023}\natexlab{}.
\newblock \bibinfo{title}{Scalable Diffusion Models with Transformers}.
\newblock
\showeprint[arxiv]{2212.09748}~[cs.CV]
\urldef\tempurl%
\url{https://arxiv.org/abs/2212.09748}
\showURL{%
\tempurl}


\bibitem[Potamias et~al\mbox{.}(2025)]%
        {potamias2025wilorendtoend3dhand}
\bibfield{author}{\bibinfo{person}{Rolandos~Alexandros Potamias}, \bibinfo{person}{Jinglei Zhang}, \bibinfo{person}{Jiankang Deng}, {and} \bibinfo{person}{Stefanos Zafeiriou}.} \bibinfo{year}{2025}\natexlab{}.
\newblock \bibinfo{title}{WiLoR: End-to-end 3D Hand Localization and Reconstruction in-the-wild}.
\newblock
\showeprint[arxiv]{2409.12259}~[cs.CV]
\urldef\tempurl%
\url{https://arxiv.org/abs/2409.12259}
\showURL{%
\tempurl}


\bibitem[Ravi et~al\mbox{.}(2024)]%
        {ravi2024sam2segmentimages}
\bibfield{author}{\bibinfo{person}{Nikhila Ravi}, \bibinfo{person}{Valentin Gabeur}, \bibinfo{person}{Yuan-Ting Hu}, \bibinfo{person}{Ronghang Hu}, \bibinfo{person}{Chaitanya Ryali}, \bibinfo{person}{Tengyu Ma}, \bibinfo{person}{Haitham Khedr}, \bibinfo{person}{Roman Rädle}, \bibinfo{person}{Chloe Rolland}, \bibinfo{person}{Laura Gustafson}, \bibinfo{person}{Eric Mintun}, \bibinfo{person}{Junting Pan}, \bibinfo{person}{Kalyan~Vasudev Alwala}, \bibinfo{person}{Nicolas Carion}, \bibinfo{person}{Chao-Yuan Wu}, \bibinfo{person}{Ross Girshick}, \bibinfo{person}{Piotr Dollár}, {and} \bibinfo{person}{Christoph Feichtenhofer}.} \bibinfo{year}{2024}\natexlab{}.
\newblock \bibinfo{title}{SAM 2: Segment Anything in Images and Videos}.
\newblock
\showeprint[arxiv]{2408.00714}~[cs.CV]
\urldef\tempurl%
\url{https://arxiv.org/abs/2408.00714}
\showURL{%
\tempurl}


\bibitem[Ren et~al\mbox{.}(2025)]%
        {ren2025motiontracksunifiedrepresentation}
\bibfield{author}{\bibinfo{person}{Juntao Ren}, \bibinfo{person}{Priya Sundaresan}, \bibinfo{person}{Dorsa Sadigh}, \bibinfo{person}{Sanjiban Choudhury}, {and} \bibinfo{person}{Jeannette Bohg}.} \bibinfo{year}{2025}\natexlab{}.
\newblock \bibinfo{title}{Motion Tracks: A Unified Representation for Human-Robot Transfer in Few-Shot Imitation Learning}.
\newblock
\showeprint[arxiv]{2501.06994}~[cs.RO]
\urldef\tempurl%
\url{https://arxiv.org/abs/2501.06994}
\showURL{%
\tempurl}


\bibitem[Romero et~al\mbox{.}(2022)]%
        {romero2022embodied}
\bibfield{author}{\bibinfo{person}{Javier Romero}, \bibinfo{person}{Dimitrios Tzionas}, {and} \bibinfo{person}{Michael~J Black}.} \bibinfo{year}{2022}\natexlab{}.
\newblock \showarticletitle{Embodied hands: Modeling and capturing hands and bodies together}.
\newblock \bibinfo{journal}{\emph{arXiv preprint arXiv:2201.02610}} (\bibinfo{year}{2022}).
\newblock


\bibitem[Shan et~al\mbox{.}(2020)]%
        {shan2020understandinghumanhandscontact}
\bibfield{author}{\bibinfo{person}{Dandan Shan}, \bibinfo{person}{Jiaqi Geng}, \bibinfo{person}{Michelle Shu}, {and} \bibinfo{person}{David~F. Fouhey}.} \bibinfo{year}{2020}\natexlab{}.
\newblock \bibinfo{title}{Understanding Human Hands in Contact at Internet Scale}.
\newblock
\showeprint[arxiv]{2006.06669}~[cs.CV]
\urldef\tempurl%
\url{https://arxiv.org/abs/2006.06669}
\showURL{%
\tempurl}


\bibitem[Shi et~al\mbox{.}(2025)]%
        {shi2025hograspflowexploringvisionbasedgenerative}
\bibfield{author}{\bibinfo{person}{Yitian Shi}, \bibinfo{person}{Zicheng Guo}, \bibinfo{person}{Rosa Wolf}, \bibinfo{person}{Edgar Welte}, {and} \bibinfo{person}{Rania Rayyes}.} \bibinfo{year}{2025}\natexlab{}.
\newblock \bibinfo{title}{HOGraspFlow: Exploring Vision-based Generative Grasp Synthesis with Hand-Object Priors and Taxonomy Awareness}.
\newblock
\showeprint[arxiv]{2509.16871}~[cs.RO]
\urldef\tempurl%
\url{https://arxiv.org/abs/2509.16871}
\showURL{%
\tempurl}


\bibitem[Sundermeyer et~al\mbox{.}(2021)]%
        {sundermeyer2021contactgraspnetefficient6dofgrasp}
\bibfield{author}{\bibinfo{person}{Martin Sundermeyer}, \bibinfo{person}{Arsalan Mousavian}, \bibinfo{person}{Rudolph Triebel}, {and} \bibinfo{person}{Dieter Fox}.} \bibinfo{year}{2021}\natexlab{}.
\newblock \bibinfo{title}{Contact-GraspNet: Efficient 6-DoF Grasp Generation in Cluttered Scenes}.
\newblock
\showeprint[arxiv]{2103.14127}~[cs.RO]
\urldef\tempurl%
\url{https://arxiv.org/abs/2103.14127}
\showURL{%
\tempurl}


\bibitem[Tang et~al\mbox{.}(2025b)]%
        {tang2025functofunctioncentriconeshotimitation}
\bibfield{author}{\bibinfo{person}{Chao Tang}, \bibinfo{person}{Anxing Xiao}, \bibinfo{person}{Yuhong Deng}, \bibinfo{person}{Tianrun Hu}, \bibinfo{person}{Wenlong Dong}, \bibinfo{person}{Hanbo Zhang}, \bibinfo{person}{David Hsu}, {and} \bibinfo{person}{Hong Zhang}.} \bibinfo{year}{2025}\natexlab{b}.
\newblock \bibinfo{title}{FUNCTO: Function-Centric One-Shot Imitation Learning for Tool Manipulation}.
\newblock
\showeprint[arxiv]{2502.11744}~[cs.RO]
\urldef\tempurl%
\url{https://arxiv.org/abs/2502.11744}
\showURL{%
\tempurl}


\bibitem[Tang et~al\mbox{.}(2025c)]%
        {tang2025mimicfuncimitatingtoolmanipulation}
\bibfield{author}{\bibinfo{person}{Chao Tang}, \bibinfo{person}{Anxing Xiao}, \bibinfo{person}{Yuhong Deng}, \bibinfo{person}{Tianrun Hu}, \bibinfo{person}{Wenlong Dong}, \bibinfo{person}{Hanbo Zhang}, \bibinfo{person}{David Hsu}, {and} \bibinfo{person}{Hong Zhang}.} \bibinfo{year}{2025}\natexlab{c}.
\newblock \bibinfo{title}{MimicFunc: Imitating Tool Manipulation from a Single Human Video via Functional Correspondence}.
\newblock
\showeprint[arxiv]{2508.13534}~[cs.RO]
\urldef\tempurl%
\url{https://arxiv.org/abs/2508.13534}
\showURL{%
\tempurl}


\bibitem[Tang et~al\mbox{.}(2025a)]%
        {tang2025trajectoryconditionedcrossembodimentskill}
\bibfield{author}{\bibinfo{person}{YuHang Tang}, \bibinfo{person}{Yixuan Lou}, \bibinfo{person}{Pengfei Han}, \bibinfo{person}{Haoming Song}, \bibinfo{person}{Xinyi Ye}, \bibinfo{person}{Dong Wang}, {and} \bibinfo{person}{Bin Zhao}.} \bibinfo{year}{2025}\natexlab{a}.
\newblock \bibinfo{title}{Trajectory Conditioned Cross-embodiment Skill Transfer}.
\newblock
\showeprint[arxiv]{2510.07773}~[cs.RO]
\urldef\tempurl%
\url{https://arxiv.org/abs/2510.07773}
\showURL{%
\tempurl}


\bibitem[Team(2025)]%
        {GalaxeaManipSim}
\bibfield{author}{\bibinfo{person}{Galaxea Team}.} \bibinfo{year}{2025}\natexlab{}.
\newblock \showarticletitle{Galaxea Manipulation Simulator}.
\newblock


\bibitem[Wang et~al\mbox{.}(2023)]%
        {wang2023mimicplaylonghorizonimitationlearning}
\bibfield{author}{\bibinfo{person}{Chen Wang}, \bibinfo{person}{Linxi Fan}, \bibinfo{person}{Jiankai Sun}, \bibinfo{person}{Ruohan Zhang}, \bibinfo{person}{Li Fei-Fei}, \bibinfo{person}{Danfei Xu}, \bibinfo{person}{Yuke Zhu}, {and} \bibinfo{person}{Anima Anandkumar}.} \bibinfo{year}{2023}\natexlab{}.
\newblock \bibinfo{title}{MimicPlay: Long-Horizon Imitation Learning by Watching Human Play}.
\newblock
\showeprint[arxiv]{2302.12422}~[cs.RO]
\urldef\tempurl%
\url{https://arxiv.org/abs/2302.12422}
\showURL{%
\tempurl}


\bibitem[Wen et~al\mbox{.}(2024)]%
        {wen2024foundationposeunified6dpose}
\bibfield{author}{\bibinfo{person}{Bowen Wen}, \bibinfo{person}{Wei Yang}, \bibinfo{person}{Jan Kautz}, {and} \bibinfo{person}{Stan Birchfield}.} \bibinfo{year}{2024}\natexlab{}.
\newblock \bibinfo{title}{FoundationPose: Unified 6D Pose Estimation and Tracking of Novel Objects}.
\newblock
\showeprint[arxiv]{2312.08344}~[cs.CV]
\urldef\tempurl%
\url{https://arxiv.org/abs/2312.08344}
\showURL{%
\tempurl}


\bibitem[Wu et~al\mbox{.}(2024)]%
        {wu2024gellogenerallowcostintuitive}
\bibfield{author}{\bibinfo{person}{Philipp Wu}, \bibinfo{person}{Yide Shentu}, \bibinfo{person}{Zhongke Yi}, \bibinfo{person}{Xingyu Lin}, {and} \bibinfo{person}{Pieter Abbeel}.} \bibinfo{year}{2024}\natexlab{}.
\newblock \bibinfo{title}{GELLO: A General, Low-Cost, and Intuitive Teleoperation Framework for Robot Manipulators}.
\newblock
\showeprint[arxiv]{2309.13037}~[cs.RO]
\urldef\tempurl%
\url{https://arxiv.org/abs/2309.13037}
\showURL{%
\tempurl}


\bibitem[Wu and Keogh(2020)]%
        {wu2020fastdtw}
\bibfield{author}{\bibinfo{person}{Renjie Wu} {and} \bibinfo{person}{Eamonn~J Keogh}.} \bibinfo{year}{2020}\natexlab{}.
\newblock \showarticletitle{FastDTW is approximate and generally slower than the algorithm it approximates}.
\newblock \bibinfo{journal}{\emph{IEEE Transactions on Knowledge and Data Engineering}} \bibinfo{volume}{34}, \bibinfo{number}{8} (\bibinfo{year}{2020}), \bibinfo{pages}{3779--3785}.
\newblock


\bibitem[Xie et~al\mbox{.}(2025)]%
        {xie2025human2robotlearningrobotactions}
\bibfield{author}{\bibinfo{person}{Sicheng Xie}, \bibinfo{person}{Haidong Cao}, \bibinfo{person}{Zejia Weng}, \bibinfo{person}{Zhen Xing}, \bibinfo{person}{Haoran Chen}, \bibinfo{person}{Shiwei Shen}, \bibinfo{person}{Jiaqi Leng}, \bibinfo{person}{Zuxuan Wu}, {and} \bibinfo{person}{Yu-Gang Jiang}.} \bibinfo{year}{2025}\natexlab{}.
\newblock \bibinfo{title}{Human2Robot: Learning Robot Actions from Paired Human-Robot Videos}.
\newblock
\showeprint[arxiv]{2502.16587}~[cs.RO]
\urldef\tempurl%
\url{https://arxiv.org/abs/2502.16587}
\showURL{%
\tempurl}


\bibitem[Xu et~al\mbox{.}(2024)]%
        {xu2024flowcrossdomainmanipulationinterface}
\bibfield{author}{\bibinfo{person}{Mengda Xu}, \bibinfo{person}{Zhenjia Xu}, \bibinfo{person}{Yinghao Xu}, \bibinfo{person}{Cheng Chi}, \bibinfo{person}{Gordon Wetzstein}, \bibinfo{person}{Manuela Veloso}, {and} \bibinfo{person}{Shuran Song}.} \bibinfo{year}{2024}\natexlab{}.
\newblock \bibinfo{title}{Flow as the Cross-Domain Manipulation Interface}.
\newblock
\showeprint[arxiv]{2407.15208}~[cs.RO]
\urldef\tempurl%
\url{https://arxiv.org/abs/2407.15208}
\showURL{%
\tempurl}


\bibitem[Yan(2025)]%
        {foundationposeplusplus}
\bibfield{author}{\bibinfo{person}{Wenhao Yan}.} \bibinfo{year}{2025}\natexlab{}.
\newblock \bibinfo{title}{FoundationPose++: Simple Tricks Boost FoundationPose Performance in High-Dynamic Scenes}.
\newblock
\urldef\tempurl%
\url{https://github.com/teal024/FoundationPose-plus-plus}
\showURL{%
\tempurl}


\bibitem[Yuan et~al\mbox{.}(2024)]%
        {yuan2024generalflowfoundationaffordance}
\bibfield{author}{\bibinfo{person}{Chengbo Yuan}, \bibinfo{person}{Chuan Wen}, \bibinfo{person}{Tong Zhang}, {and} \bibinfo{person}{Yang Gao}.} \bibinfo{year}{2024}\natexlab{}.
\newblock \bibinfo{title}{General Flow as Foundation Affordance for Scalable Robot Learning}.
\newblock
\showeprint[arxiv]{2401.11439}~[cs.RO]
\urldef\tempurl%
\url{https://arxiv.org/abs/2401.11439}
\showURL{%
\tempurl}


\bibitem[Yuan et~al\mbox{.}(2025)]%
        {yuan2025demograspuniversaldexterousgrasping}
\bibfield{author}{\bibinfo{person}{Haoqi Yuan}, \bibinfo{person}{Ziye Huang}, \bibinfo{person}{Ye Wang}, \bibinfo{person}{Chuan Mao}, \bibinfo{person}{Chaoyi Xu}, {and} \bibinfo{person}{Zongqing Lu}.} \bibinfo{year}{2025}\natexlab{}.
\newblock \bibinfo{title}{DemoGrasp: Universal Dexterous Grasping from a Single Demonstration}.
\newblock
\showeprint[arxiv]{2509.22149}~[cs.RO]
\urldef\tempurl%
\url{https://arxiv.org/abs/2509.22149}
\showURL{%
\tempurl}


\bibitem[Zhao et~al\mbox{.}(2021)]%
        {zhao2021pointtransformer}
\bibfield{author}{\bibinfo{person}{Hengshuang Zhao}, \bibinfo{person}{Li Jiang}, \bibinfo{person}{Jiaya Jia}, \bibinfo{person}{Philip Torr}, {and} \bibinfo{person}{Vladlen Koltun}.} \bibinfo{year}{2021}\natexlab{}.
\newblock \bibinfo{title}{Point Transformer}.
\newblock
\showeprint[arxiv]{2012.09164}~[cs.CV]
\urldef\tempurl%
\url{https://arxiv.org/abs/2012.09164}
\showURL{%
\tempurl}


\bibitem[Zhao et~al\mbox{.}(2025)]%
        {zhao2025tasterobadvancingvideogeneration}
\bibfield{author}{\bibinfo{person}{Hongxiang Zhao}, \bibinfo{person}{Xingchen Liu}, \bibinfo{person}{Mutian Xu}, \bibinfo{person}{Yiming Hao}, \bibinfo{person}{Weikai Chen}, {and} \bibinfo{person}{Xiaoguang Han}.} \bibinfo{year}{2025}\natexlab{}.
\newblock \bibinfo{title}{TASTE-Rob: Advancing Video Generation of Task-Oriented Hand-Object Interaction for Generalizable Robotic Manipulation}.
\newblock
\showeprint[arxiv]{2503.11423}~[cs.CV]
\urldef\tempurl%
\url{https://arxiv.org/abs/2503.11423}
\showURL{%
\tempurl}


\bibitem[Zhao et~al\mbox{.}(2023)]%
        {zhao2023learningfinegrainedbimanualmanipulation}
\bibfield{author}{\bibinfo{person}{Tony~Z. Zhao}, \bibinfo{person}{Vikash Kumar}, \bibinfo{person}{Sergey Levine}, {and} \bibinfo{person}{Chelsea Finn}.} \bibinfo{year}{2023}\natexlab{}.
\newblock \bibinfo{title}{Learning Fine-Grained Bimanual Manipulation with Low-Cost Hardware}.
\newblock
\showeprint[arxiv]{2304.13705}~[cs.RO]
\urldef\tempurl%
\url{https://arxiv.org/abs/2304.13705}
\showURL{%
\tempurl}


\bibitem[Zhaxizhuoma et~al\mbox{.}(2025)]%
        {zhaxizhuoma2025fastumiscalablehardwareindependentuniversal}
\bibfield{author}{\bibinfo{person}{Zhaxizhuoma}, \bibinfo{person}{Kehui Liu}, \bibinfo{person}{Chuyue Guan}, \bibinfo{person}{Zhongjie Jia}, \bibinfo{person}{Ziniu Wu}, \bibinfo{person}{Xin Liu}, \bibinfo{person}{Tianyu Wang}, \bibinfo{person}{Shuai Liang}, \bibinfo{person}{Pengan Chen}, \bibinfo{person}{Pingrui Zhang}, \bibinfo{person}{Haoming Song}, \bibinfo{person}{Delin Qu}, \bibinfo{person}{Dong Wang}, \bibinfo{person}{Zhigang Wang}, \bibinfo{person}{Nieqing Cao}, \bibinfo{person}{Yan Ding}, \bibinfo{person}{Bin Zhao}, {and} \bibinfo{person}{Xuelong Li}.} \bibinfo{year}{2025}\natexlab{}.
\newblock \bibinfo{title}{FastUMI: A Scalable and Hardware-Independent Universal Manipulation Interface with Dataset}.
\newblock
\showeprint[arxiv]{2409.19499}~[cs.RO]
\urldef\tempurl%
\url{https://arxiv.org/abs/2409.19499}
\showURL{%
\tempurl}


\bibitem[Zhi et~al\mbox{.}(2025)]%
        {zhi20253dflowactionlearningcrossembodimentmanipulation}
\bibfield{author}{\bibinfo{person}{Hongyan Zhi}, \bibinfo{person}{Peihao Chen}, \bibinfo{person}{Siyuan Zhou}, \bibinfo{person}{Yubo Dong}, \bibinfo{person}{Quanxi Wu}, \bibinfo{person}{Lei Han}, {and} \bibinfo{person}{Mingkui Tan}.} \bibinfo{year}{2025}\natexlab{}.
\newblock \bibinfo{title}{3DFlowAction: Learning Cross-Embodiment Manipulation from 3D Flow World Model}.
\newblock
\showeprint[arxiv]{2506.06199}~[cs.RO]
\urldef\tempurl%
\url{https://arxiv.org/abs/2506.06199}
\showURL{%
\tempurl}


\bibitem[Zhou et~al\mbox{.}(2025)]%
        {zhou2025teachoncelearnoneshot}
\bibfield{author}{\bibinfo{person}{Huayi Zhou}, \bibinfo{person}{Ruixiang Wang}, \bibinfo{person}{Yunxin Tai}, \bibinfo{person}{Yueci Deng}, \bibinfo{person}{Guiliang Liu}, {and} \bibinfo{person}{Kui Jia}.} \bibinfo{year}{2025}\natexlab{}.
\newblock \bibinfo{title}{You Only Teach Once: Learn One-Shot Bimanual Robotic Manipulation from Video Demonstrations}.
\newblock
\showeprint[arxiv]{2501.14208}~[cs.RO]
\urldef\tempurl%
\url{https://arxiv.org/abs/2501.14208}
\showURL{%
\tempurl}


\bibitem[Zhu et~al\mbox{.}(2025)]%
        {zhu2025touchwildlearningfinegrained}
\bibfield{author}{\bibinfo{person}{Xinyue Zhu}, \bibinfo{person}{Binghao Huang}, {and} \bibinfo{person}{Yunzhu Li}.} \bibinfo{year}{2025}\natexlab{}.
\newblock \bibinfo{title}{Touch in the Wild: Learning Fine-Grained Manipulation with a Portable Visuo-Tactile Gripper}.
\newblock
\showeprint[arxiv]{2507.15062}~[cs.RO]
\urldef\tempurl%
\url{https://arxiv.org/abs/2507.15062}
\showURL{%
\tempurl}


\bibitem[Zhu et~al\mbox{.}(2020)]%
        {zhu2020robosuitemodularsimulationframework}
\bibfield{author}{\bibinfo{person}{Yuke Zhu}, \bibinfo{person}{Josiah Wong}, \bibinfo{person}{Ajay Mandlekar}, \bibinfo{person}{Roberto Martín-Martín}, \bibinfo{person}{Abhishek Joshi}, \bibinfo{person}{Soroush Nasiriany}, {and} \bibinfo{person}{Yifeng Zhu}.} \bibinfo{year}{2020}\natexlab{}.
\newblock \bibinfo{title}{robosuite: A Modular Simulation Framework and Benchmark for Robot Learning}.
\newblock
\showeprint[arxiv]{2009.12293}~[cs.RO]
\urldef\tempurl%
\url{https://arxiv.org/abs/2009.12293}
\showURL{%
\tempurl}


\end{thebibliography}
